\documentclass{article}

\usepackage{microtype}
\usepackage{graphicx}
\usepackage{subcaption}
\usepackage{booktabs} %

\usepackage{hyperref}

\usepackage[accepted]{icml2026}

\usepackage{amsmath}
\usepackage{amssymb}
\usepackage{mathtools}
\usepackage{amsthm}

\usepackage[capitalize,noabbrev]{cleveref}

\theoremstyle{plain}

\theoremstyle{definition}

\theoremstyle{remark}

\icmltitlerunning{Reasoning Structure of Large Language Models}

\usepackage[utf8]{inputenc}
\usepackage{tabularx} %
\newcolumntype{Y}{>{\centering\arraybackslash}X}
\newcolumntype{C}[1]{>{\centering\arraybackslash}p{#1}}
\usepackage{multirow}
\usepackage{nicefrac}
\usepackage{bbding} %

\usepackage{pifont}

\usepackage{mdframed}
\usepackage{verbatim}
\usepackage{fvextra}
\newcommand{\displayprompt}[1]{%
  \begin{mdframed}[backgroundcolor=gray!10, linecolor=black]
    \VerbatimInput[
      breaklines=true,
      breakanywhere=true,
      fontsize=\tiny,
      baselinestretch=0.2,
      breaksymbolleft=
    ]{#1}
  \end{mdframed}
}

\usepackage{tikz}
\usetikzlibrary{shapes.geometric, arrows.meta, positioning, fit, calc}

\begin{document}

\twocolumn[
  \icmltitle{Reasoning Structure of Large Language Models}

  \icmlsetsymbol{equal}{*}
  \begin{icmlauthorlist}
    \icmlauthor{Frédéric Berdoz}{eth}
    \icmlauthor{Luca A. Lanzendörfer}{eth}
    \icmlauthor{Fabian Farestam}{eth}
    \icmlauthor{Roger Wattenhofer}{eth}
  \end{icmlauthorlist}
  \icmlaffiliation{eth}{ETH Zurich, Switzerland}
  \icmlcorrespondingauthor{Frédéric Berdoz}{fberdoz@ethz.ch}

  \icmlkeywords{Machine Learning, ICML}

  \vskip 0.3in
]

\printAffiliationsAndNotice{Code available at \url{https://github.com/ETH-DISCO/llm-reasoning-efficiency}.}

\begin{abstract}
Large reasoning models (LRMs) are often evaluated using metrics such as final-answer accuracy or token count. However, identical scores on these metrics can hide fundamentally different reasoning structures.
To address this limitation, we introduce a scalable LRM benchmark of logic puzzles and a pipeline that converts unstructured traces into verifiable reasoning graphs of claims and dependencies. This turns reasoning into a structured, measurable object whose topology can be quantitatively analyzed.
Building on this, we define a reasoning efficiency metric that quantifies how concentrated the model's logical flow is. Our analysis on open-source reasoning models shows that structural measurements separate behaviors that token count and accuracy conflate, providing a practical tool for diagnosing failure modes and comparing how reasoning scales with puzzle difficulty.
\end{abstract}

\begin{figure}[t]
\centering
\begin{minipage}[t]{0.49\columnwidth}
    \centering
    \includegraphics[width=\linewidth]{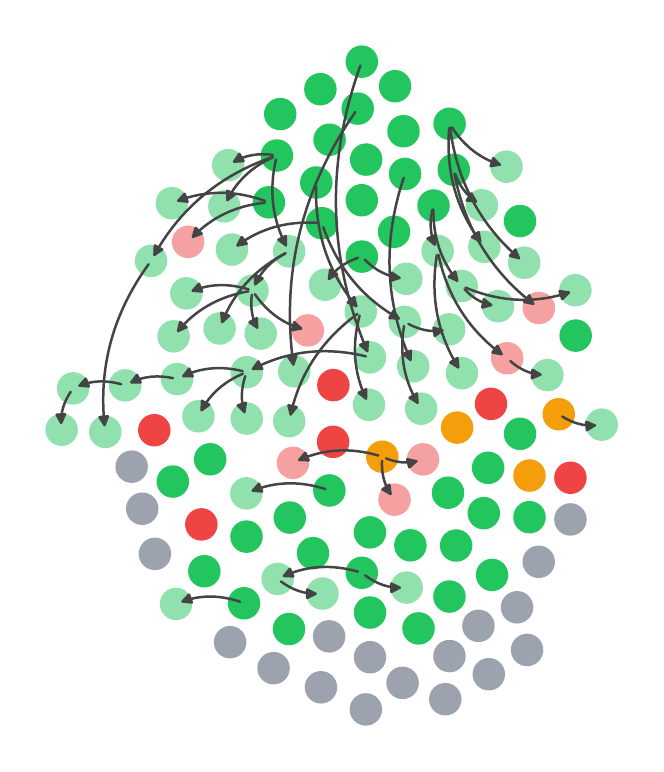}
    $\eta = 0.059$
\end{minipage}%
\begin{minipage}[t]{0.49\columnwidth}
    \centering
    \includegraphics[trim = 0 -100 0 0, width=0.8\linewidth]{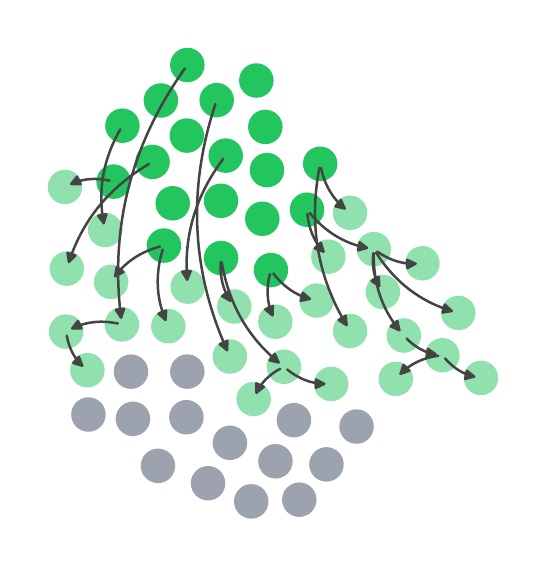}
    $\eta = 0.362$
\end{minipage}
\caption{\textbf{Qualitative reasoning graphs.} Shown are two extracted claim graphs from two independent samples of the same model (Qwen3 235B) on the same \emph{Tents} puzzle instance: a diffuse trace (left) and a focused trace (right). Both traces solve the instance. Nodes are atomic claims and edges indicate deductive dependencies. Node color encodes claim validity: green = verified correct, red = verified wrong, grey = unverifiable, orange = tentative. Node opacity encodes claim status: opaque = stated, semi-transparent = derived. Accuracy and token count alone are too coarse to provide meaningful feedback about how the model reasoned, whereas graph structure and efficiency (annotated $\eta$) distinguish diffuse exploration from focused deductions.}
\label{fig:graph_examples}
\end{figure}

\section{Introduction}

Reasoning is central to human intelligence and remains a major challenge for machine learning systems.
Thanks to their ability to exploit increased test-time compute through long Chain-of-Thought (CoT) traces \cite{wei2022chain}, Large Reasoning Models (LRMs) have shown impressive performance on a broad set of reasoning tasks, including complex coding \cite{chen2021evaluating}, logical deduction \cite{lin2025zebralogic}, mathematical reasoning \cite{cobbe2021training}, and spatial reasoning \cite{berdoz2026text}.
However, because most evaluations collapse behavior into one-dimensional metrics such as final-answer accuracy or token count, it remains unclear \emph{how} these models reason.
This gap has motivated prior work to develop controllable puzzle environments that are less prone to benchmark saturation and data contamination \cite{chen2025enigmata, zhang2025puzzlebench}.
Logic puzzles have long attracted human curiosity because they are \textit{``easy to learn, but hard to master.''} Unlike many real-world tasks, they are fully specified and admit unambiguous verification. Their difficulty can be scaled without changing the underlying rules, making them an ideal controlled benchmark for studying reasoning behavior.
For example, \citet{shojaee2025illusion} analyzed four puzzle families with adjustable complexity and found sharp regime changes and eventual collapse beyond a critical difficulty threshold.
The authors also reported trace-level phenomena including overthinking, early fixation on incorrect hypotheses, and counterintuitive reductions in ``reasoning effort'' near failure \cite{shojaee2025illusion}.
Together, these results suggest that controllable puzzles can reveal failure modes that accuracy alone misses, and motivate analyses that use the intermediate traces rather than only the final answer.

\begin{figure*}[t]
    \centering
    \input{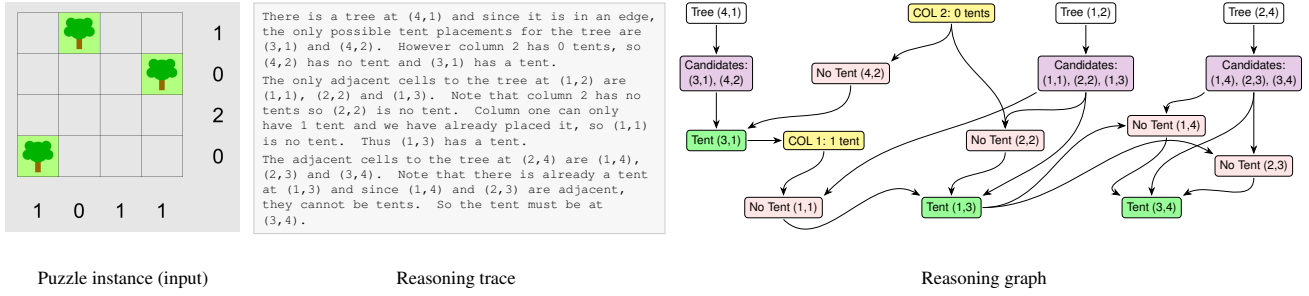}
    \caption{\textbf{Reasoning graph extraction.} Example of a 4$\times$4 \emph{Tents} instance (left), an excerpt of a \emph{human-generated} reasoning trace (middle), and the corresponding verifiable claim graph extracted using our pipeline  described in \cref{sec:methodology} (right). This illustrates how our pipeline turns free-form traces into structured objects that can be analyzed beyond accuracy and token count.}
    \label{fig:benchmark_overview}
\end{figure*}

In this work, we move from measuring \emph{how much} a model thinks to measuring the \emph{structure} of its reasoning.
We introduce a scalable benchmark of 2D grid puzzles and convert free-form textual traces into verifiable reasoning graphs of claims and dependencies, turning reasoning into a structured object that supports quantitative analysis.
Graph topology, including depth and the fraction of the graph that supports the final solution, captures differences between focused reasoning and diffuse exploration that token count and final accuracy fail to capture.
Building on this representation, we define a reasoning-flow efficiency metric $\eta$ that measures the concentration of the model’s logical flow between its observations and its proposed solution. This metric can differentiate reasoning traces even when they reach the same
correct solution with similar trace length. As \cref{fig:efficiency_scaling} shows, $\eta$ exposes differences in reasoning structure between models even in regimes where accuracy is saturated and token budgets overlap.

Analyzing reasoning traces, we find that token count is a suboptimal proxy for reasoning quality, since extra tokens primarily translate into verification overhead. In comparison, our proposed metric is able to track solution-relevant structure and correctness more effectively. Furthermore, our benchmark shows that even as models spend considerably more tokens at higher difficulty levels the hardest regime remains largely unsolved, indicating current limitations that are not resolved by simply allocating more compute.

We summarize our contributions as follows:
\begin{itemize}
    \item We introduce a scalable benchmark of 21 logic puzzles that enables controlled scaling studies of LLM reasoning.
    \item We convert free-form textual reasoning traces into verifiable reasoning graphs of claims and dependencies, making reasoning topology measurable beyond accuracy and token count metrics.
    \item We define a reasoning efficiency metric $\eta$ that separates focused reasoning from diffuse exploration by measuring how concentrated logical flow is relative to the minimal claim set.
\end{itemize}

\section{Related Work}

\subsection{Eliciting Reasoning in LLMs}
\paragraph{Prompting, search, and RL at test time.}
Chain-of-Thought prompting elicits intermediate reasoning steps in language models \cite{wei2022chain}. Follow-up work adds explicit test-time search, including Tree-of-Thoughts with branching and backtracking \cite{yao2023tree} and Graph-of-Thoughts \cite{besta2024graph}, with adaptive variants that build task-specific DAG decompositions \cite{pandey2025adaptive}. In parallel, reinforcement learning with verifiable rewards (RLVR) uses verifiable signals (e.g., unit tests or environment validators) to improve deliberation while reducing reward hacking concerns \cite{lambert2024tulu3,amodei2016concrete}, and has enabled strong reasoning performance in recent systems \cite{openai2024o1,deepseek2025r1}. Rather than proposing new elicitation mechanisms, we analyze the structures these approaches produce by reconstructing verifiable trees from traces and studying how topology changes with puzzle complexity.

\subsection{Benchmarking Reasoning}
\begin{figure*}[t]
\centering
\includegraphics[width=\textwidth]{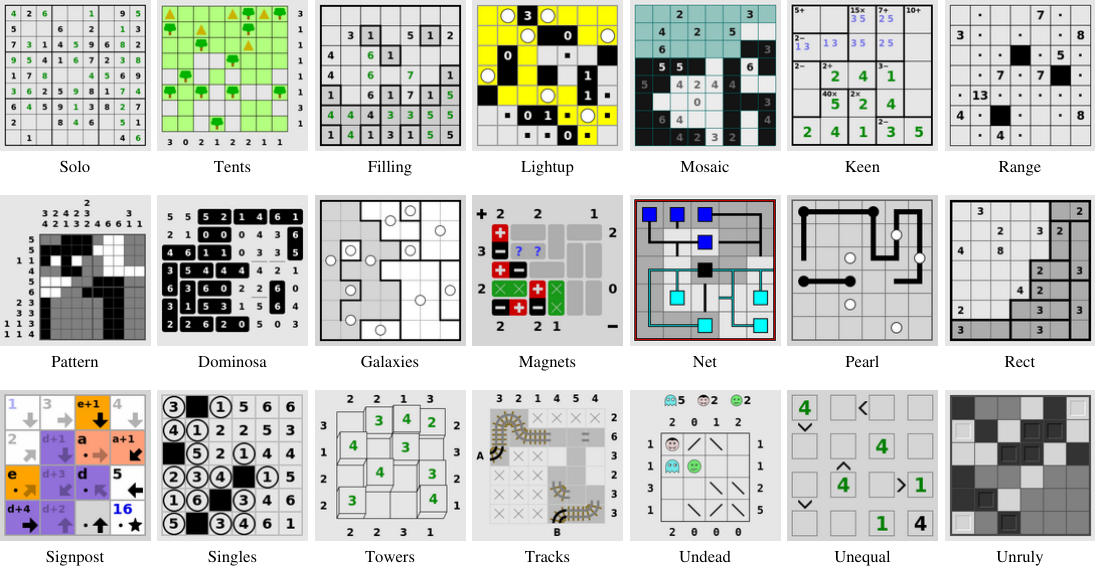}
\caption{\textbf{Puzzle suite overview.} We evaluate models on 21 grid puzzles spanning diverse constraint types (e.g., placement, connectivity, counting, and Latin-square-style constraints), each with four difficulty levels (\emph{Trivial}, \emph{Human easy}, \emph{Human normal}, \emph{Human hard}).. Detailed rules, state representations, and difficulty-to-size mappings for each puzzle are provided in \cref{app:puzzle_descriptions}. The puzzle images are taken from \citet{tatham2025portable}.}
\label{fig:puzzle_grid}
\end{figure*}

\paragraph{Textual and mathematical reasoning.}
Many benchmarks evaluate final answers on static text without an executable intermediate state. ProofWriter, ProntoQA, and LogicBench require multi-step inference but express intermediate steps in free-form language, which prevents step-level verification against an environment \cite{tafjord2021proofwriter,saparov2023language,parmar2024logicbench}. SATBench validates final solutions with solvers, but remains text-centric rather than operating within a manipulable state machine \cite{wei2025satbench}. Math benchmarks such as GSM8K emphasize multi-step computation to a final output, but do not provide an environment with executable actions and verifiable state transitions \cite{cobbe2021training}.

\paragraph{Controllable logical reasoning benchmarks.}
As widely used benchmarks saturate, recent evaluations scale difficulty in controlled ways. GSM-Symbolic shows brittleness under small numerical changes or irrelevant clauses \cite{mirzadeh2025gsm}. ZebraLogic uses scalable logic-grid puzzles and identifies a curse of complexity as constraints increase \cite{lin2025zebralogic}. SATBench generates puzzles from SAT formulas and highlights hard-UNSAT and search-related failure modes \cite{wei2025satbench}. While these settings typically focus on outcome metrics (e.g., accuracy, pass rates), we add structural measurements over reconstructed reasoning trees.

\paragraph{Interactive games and agent evaluations.}
Interactive benchmarks move toward agentic evaluation, but often lack a shared, verifiable state semantics across tasks. SmartPlay spans diverse games, which complicates defining a canonical state representation and move semantics for step-level verification \cite{wu2024smartplay}. PuzzleBench generates dynamic visual puzzle instances with verification, but its focus is multimodal perception rather than explicit state-transition systems for search and planning \cite{zhang2025puzzlebench}.

\paragraph{Puzzle environments with verifiability.}
Puzzle benchmarks emphasize verifiability and scalable difficulty, but differ in what is verified. Enigmata provides generator-verifier pairs for scalable evaluation and RLVR-style training, but verifies only final submissions \cite{chen2025enigmata}. \citet{shojaee2025illusion} study trace-level collapse under increasing complexity in parameterized puzzle settings, but focus on a small set of puzzle types. 
\subsection{Studying Reasoning}
\paragraph{Behavioral studies.}
Recent work characterizes failure modes of long-trace reasoning beyond final-answer accuracy, including overthinking on trivial problems \cite{chen2025do}, underthinking via premature switching between lines of thought \cite{wang2025thoughts}, diversity collapse where pass@1 improves while pass@k degrades \cite{dang2025assessing}, and missing-premise overthinking on ill-posed questions \cite{fan2025missing}. In this work, we study the structure of reasoning rather than its failure modes.

\paragraph{Trace representations.}
Structured trace representations support analysis and test-time selection. Landscape-style embeddings visualize convergence and provide verification signals \cite{zhou2025landscape}. Reasoning-graph pipelines compress traces into step graphs and link graph properties to accuracy and prompting sensitivity \cite{xiong2025mapping}, and other work derives graphs from hidden-state representations to study topological properties across models \cite{minegishi2025topology}. Related approaches propose metrics and structures to separate useful reasoning from failed exploration, including failed-step fraction for reranking and editing \cite{feng2025what}, task-specific DAG structures and consistency metrics for math reasoning \cite{zhang2025dag}, and tree-jump representations that quantify exploration and backtracking for test-time selection \cite{zeng2025rejump}. In contrast, our representation is grounded in an executable puzzle environment. We extract verifiable claims and dependencies: claim nodes are verified deterministically against the executable environment, while edges are attributed through constrained, puzzle-specific LLM rule application and validated by manual inspection (\cref{app:verification}). This grounding enables structural metrics that remain comparable across puzzle families and difficulty levels.

\section{Methodology}
\label{sec:methodology}

We present a method for analyzing the reasoning behavior of large reasoning models (LRMs) by constructing reasoning graphs from model-generated traces. These graphs enable a unified evaluation of reasoning accuracy, structural properties, and efficiency beyond final-answer correctness and token count.

\begin{figure}[t]
\centering
\newcommand{\FigFont}{\small}
\newcommand{\ArrowTip}{Latex}
\newcommand{\ArrowWidth}{0.45pt}
\newcommand{\ArrowColor}{black!60}

\newcommand{\BoxCornerRadius}{3pt}
\newcommand{\BoxInnerSep}{5pt}
\newcommand{\DotInnerSep}{0.9pt}
\newcommand{\BoxLineWidth}{0.4pt}
\newcommand{\BoxDrawColor}{black!25}
\newcommand{\DotColor}{black!50}
\newcommand{\BoxMinWidth}{38.4mm}

\definecolor{cInput}{HTML}{DBEAFE}      %
\definecolor{cProcess}{HTML}{E0F2FE}    %
\definecolor{cLLM}{HTML}{D1FAE5}        %
\definecolor{cMerge}{HTML}{FEF3C7}      %
\definecolor{cOutput}{HTML}{EDE9FE}     %

\newcommand{\NodeDistance}{6mm}
\newcommand{\BranchVerticalSep}{8mm}
\newcommand{\BranchHorizontalShift}{22.4mm}
\newcommand{\JunctionToMergeSep}{5mm}

\begin{tikzpicture}[
  font=\FigFont,
  >=\ArrowTip,
  node distance=\NodeDistance,
  box/.style={
    draw=\BoxDrawColor,
    line width=\BoxLineWidth,
    rounded corners=\BoxCornerRadius,
    align=center,
    inner sep=\BoxInnerSep,
    minimum width=\BoxMinWidth,
  },
  input/.style={box, fill=none},
  process/.style={box, fill=cProcess},
  llm/.style={box, fill=cLLM},
  merge/.style={box, fill=cMerge},
  output/.style={box, fill=none},
  dot/.style={
    circle,
    draw=\DotColor,
    fill=\DotColor,
    inner sep=\DotInnerSep
  },
  arrow/.style={
    ->,
    draw=\ArrowColor,
    line width=\ArrowWidth,
    rounded corners=3pt,
  }
]

\node[input] (trace) {Trace $S$};
\node[process, right=of trace] (chunk) {Chunking and\\pre-processing (\cref{prompt:coordinate_prompt})};

\node[process, below=8mm of trace] (det) {Deterministic\\extractor};
\node[llm] (llm_ext) at (chunk |- det) {LLM extract\\(\cref{prompt:claim_extraction})};

\node[llm, below=of det] (aug) {LLM augmentation\\(\cref{prompt:claim_complement})};
\node[merge] (mergenode) at (chunk |- aug) {Merge claims and\\LLM screen (\cref{prompt:claim_cleaning})};

\node[output, below=of aug] (claim) {Final claim set $V$};
\node[llm] (verification) at (chunk |- claim) {Claim verification};

\node[llm, below=of claim] (rule) {Rule extraction\\(\cref{prompt:rule_extraction})};
\node[output] (final) at (chunk |- rule) {Final edge set $E$};

\draw[arrow] (trace.east) -- (chunk.west);
\draw[arrow] (chunk.south) -- (llm_ext.north);
\draw[arrow] (chunk.south) -- ++(0,-3mm) -| (det.north);
\draw[arrow] (det.south) -- (aug.north);
\draw[arrow] (aug.east) -- (mergenode.west);
\draw[arrow] (llm_ext.south) -- (mergenode.north);
\draw[arrow] (mergenode.south) -- ++(0,-3mm) -| (claim.north);
\draw[arrow] (mergenode.south) -- (verification.north);
\draw[arrow] (claim.south) -- (rule.north);
\draw[arrow] (rule.east) -- (final.west);

\end{tikzpicture}
\caption{\textbf{Graph extraction overview.} Chunk-level extraction uses deterministic extraction with LLM augmentation and from-scratch LLM extraction. Outputs are merged and screened in batches. The final claim set is then processed to extract the edges that representing the canonical reasoning steps. LLM-based steps are visualized in green, pre-processing steps are shown in blue.}
\label{fig:graph_extraction_process}
\end{figure}
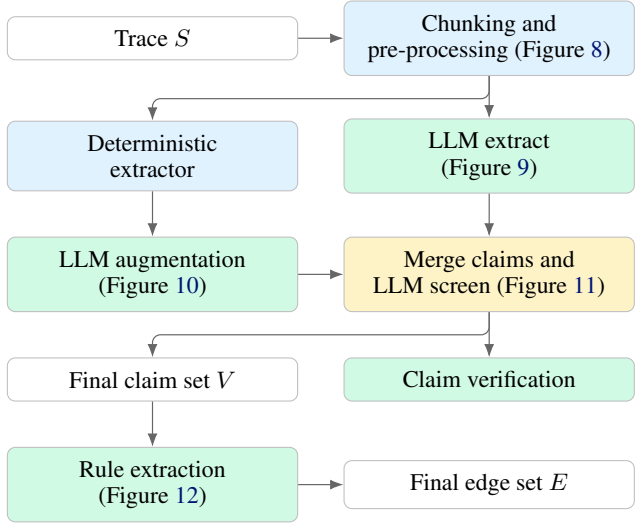

\subsection{Solving Grid Puzzles with LRMs}
We build our reasoning benchmark for LRMs on a scalable grid-based puzzle RL environment \cite{estermann2024puzzles} derived from Simon Tatham’s puzzle collection \cite{tatham2025portable}. These deterministic puzzles require abstraction, planning, and multi-step logical reasoning while remaining fully specified by explicit rules and state transitions. The executable environment further allows us to validate intermediate claims extracted from reasoning traces, rather than only final answers, enabling fine-grained analysis of both correctness and failure modes in the LRM's reasoning processes. Details about the prompting strategy to solve the puzzles are provided in \cref{app:solver_prompt}.

\subsection{From Textual Traces to Reasoning Graphs}
Our goal is to assess the quality of LLM reasoning beyond downstream accuracy or token-level metrics. To this end, we extract the latent reasoning structure underlying the textual trace produced by a reasoning model on a given puzzle instance. We represent this reasoning process as a directed graph, where nodes correspond to claims and edges capture deductive dependencies. In this section, we formalize the notation and structural assumptions used to define these graphs.

\paragraph{Atomic Claims.}
Let $S = (s_1, s_2, \dots)$ denote the reasoning trace of a LRM on a puzzle instance, represented as an ordered sequence of sentences appearing in the generated output, and let $\mathcal{C}$ be the set of all atomic claims that can be made about a puzzle. 
Examples of such claims could be: ``The cell at row 3 and column 4 is 9"  for \emph{Sudoku} (whose correctness only depends on the full solution), or ``There is already a 4 in column 5" (whose correctness depends on the current partial solution). Note that simply stating a solution for a $9\times 9$ \emph{Sudoku} requires 81 claims and that a single sentence may contain multiple distinct claims due to the free form nature of text.

\begin{table*}[t]
    \centering
    \small
    \setlength{\tabcolsep}{4pt}
    \renewcommand{\arraystretch}{0.9}
    \caption{\textbf{PUZZLE benchmark results by difficulty.} For each difficulty level we report accuracy (Acc., \%) and mean completion token count (Tok.). The final columns report mean accuracy and mean token count averaged across the four difficulty levels (Avg Acc., Avg Tok.).}
    \label{tab:benchmark_accuracy}
    \begin{tabularx}{\textwidth}{l *{10}{>{\centering\arraybackslash}X}}
    \toprule
    \multirow{2}{*}{Model}
    & \multicolumn{2}{c}{Trivial}
    & \multicolumn{2}{c}{Human easy}
    & \multicolumn{2}{c}{Human normal}
    & \multicolumn{2}{c}{Human hard}
    & \multicolumn{2}{c}{Avg} \\
    \cmidrule(lr){2-3} \cmidrule(lr){4-5} \cmidrule(lr){6-7} \cmidrule(lr){8-9} \cmidrule(lr){10-11}
    & Acc. & Tok.
    & Acc. & Tok.
    & Acc. & Tok.
    & Acc. & Tok.
    & Acc. & Tok. \\
    \midrule
    \textbf{GPT-5}      & 83.8 & 4153.5 & 69.5 & 10179.6 & 58.1 & 17273.6 & 5.7 & 19861.9 & 54.3 & 12867.2 \\
    \midrule
    \textbf{Qwen 3 235B}    & 69.5 & 10257.3 & 44.8 & 19033.0 & 21.0 & 23104.0 & 0.0 & 23608.6 & 33.8 & 19000.7 \\
    \textbf{DeepSeek V3.2}  & 77.1 & 7694.7 & 53.3 & 20632.6 & 44.8 & 27037.7 & 0.0 & 36787.4 & 43.8 & 23038.1 \\
    \textbf{Kimi K2}    & 77.1 & 10601.5 & 56.2 & 29713.7 & 41.0 & 43751.2 & 1.0 & 61307.1 & 43.8 & 36343.4 \\
    \bottomrule
    \end{tabularx}
\end{table*}

\paragraph{Reasoning Graphs.} 
Let $G=(V,E)$ be a directed graph corresponding to the reasoning trace of a LRM, where $V\subseteq \mathcal{C}\times \mathbb{N}$ denotes the set of claim occurrences extracted from $S$ (each vertex $v=(c,i)\in V$ indicates that claim $c$ is asserted in sentence $s_i$). By construction, each element in $V$ is unique: the same claim cannot appear twice in the same sentence.
Edges in $G$ represent either an inference from premises to a conclusion, or a restatement link between two occurrences of the same underlying claim. When an inference requires multiple premises, separate edges connect each premise to the derived claim. All premises of an inference appear before its conclusion in $S$, which implies a topological order over $V$. As a consequence, $G$ and all its subgraphs are directed acyclic graphs (DAGs). We also merge restated static claims into the first occurrence for graph metrics.

\paragraph{Minimal Claim Set.} For a given puzzle instance, let $C^*\subseteq \mathcal{C}$ denote the minimal set of claims that fully determine the puzzle solution. Formally, each $c \in C^*$ is a claim that is required to reconstruct the known solution and $C^*$ is minimal in the sense that removing any $c \in C^*$ would make it impossible to recover the full solution from the remaining claims without additional inference steps. In other words, $C^*$ is simply the formatted statement of the solution and we have that $C^* \subseteq \{ c\in \mathcal{C} \mid \exists (c, s) \in V \text{ for at least one } s\in S\}$ if the LRM solves the puzzle. The converse is not true however, since the LRM might give a correct solution somewhere in its reasoning trace but submit a different solution. We define $V^*_\text{sol} = \{(c, s_i) \in V \mid c\in C^*, i = \min\{j \mid (c, s_j)\in V\}\}$ 
as the set of first occurrences of claims that belong to the solution.

\paragraph{Reasoning Subgraphs for Solved Puzzles.}
We further identify two subgraphs for solved puzzle instances. First, we define the minimal solution subgraph $G_\text{sol} = (V_\text{sol}, E_\text{sol})$ of $G$ as the subgraph containing all vertices that directly or indirectly contribute to  $V^*_\text{sol}$, formally defined as follows:
\[
V_{\rm sol} = V^*_\text{sol} \cup \{\, v \in V \mid \exists\, u \in V^*_\text{sol} : v \leadsto_{\scriptscriptstyle G} u \,\},
\]
\[
E_\text{sol} = \{ (u,v) \in E \mid u,v \in V_\text{sol} \}.
\]
We use the notation $u \leadsto_{\scriptscriptstyle G} v$ to indicate that there exists a directed path from $u$ to $v$ in $G$. Similarly, let $G_{\rm ver} = (V_{\rm ver}, E_{\rm ver})$ be the verification subgraph, with 
$$
V'_{\rm ver}= V^*_\text{sol} \cup \{ v \in V \setminus  V_\text{sol} \mid \exists\, u \in V^*_\text{sol} : u \leadsto_{\scriptscriptstyle G} v  \,\}
$$ 
$$ 
V_{\rm ver}= V'_{\rm ver} \cup \{ v \in V \setminus  V_\text{sol} \mid \exists\, u \in V'_{\rm ver} : v \leadsto_{\scriptscriptstyle G} u  \,\}
$$
$$
E_{\rm ver} = \{ (u,v) \in E \mid u,v \in V_{\rm ver} \}
$$
Intuitively, $V_\text{ver}$ contains all descendants of $V^*_\text{sol}$ and all their ancestors that are not in $V_\text{sol}$.

\begin{figure*}[t]
\centering
\includegraphics[width=0.95\textwidth]{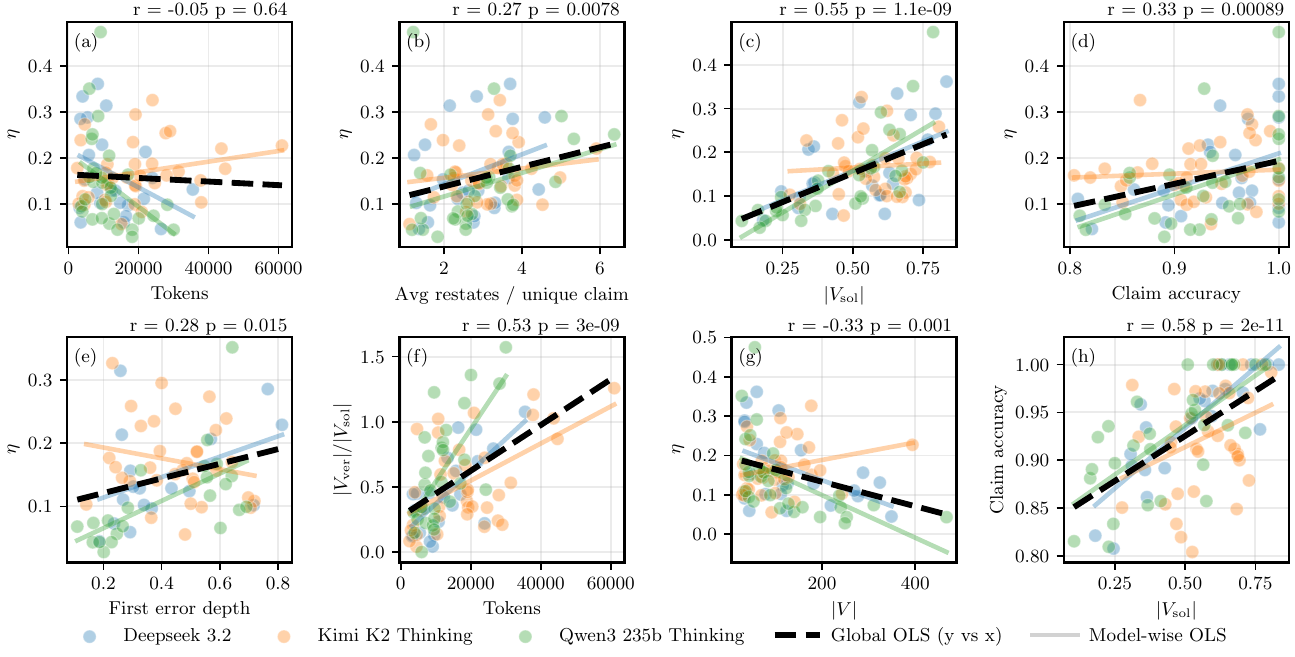}
\caption{\textbf{Efficiency correlations.} Each panel plots a graph-level metric against efficiency $\eta$ or token count. Points are colored by model; dashed lines show linear fits. Statistical significance is reported above each plot. (a) Efficiency is not significantly correlated with verbosity. (b,c) Efficiency tracks claim composition: negatively with stated fraction and positively with restated fraction. This suggests that LRM reason in a more strucured manner when they restated claims rather than stating new ones.  (d) Higher efficiency is associated with higher claim accuracy. (e) Higher efficiency is associated with later first-error depth. (f) Verification overhead ($|V_{\mathrm{ver}}|/|V_{\mathrm{sol}}|$) grows with token count.}
\label{fig:efficiency_correlations}
\end{figure*}

\subsection{Reasoning Metrics}

\paragraph{Graph-based Reasoning Metrics.}
We report graph-size and graph-topology statistics on the extracted claim graph $G=(V,E)$ and its subgraphs $G_{\mathrm{sol}}$ and $G_{\mathrm{ver}}$. We use $|V|$, $|V_{\mathrm{sol}}|$, and $|V_{\mathrm{ver}}|$ to denote the number of claim nodes in each graph. 
We define the depth of a node $v$ as the length of the longest directed path from any stated node to $v$, and the depth of the graph as the maximum depth of all nodes.
For solved instances, we quantify (i) how much of the extracted graph supports the minimal solution subgraph ($\nicefrac{|V_{\mathrm{sol}}|}{|V|}$), and (ii) how much additional reasoning is devoted to verification ($\nicefrac{|V_{\mathrm{ver}}|}{|V_{\mathrm{sol}}|}$).

\paragraph{Validity-based Reasoning Metrics.}
Each extracted verifiable claim is checked against the executable puzzle environment (see \cref{app:verification}), yielding a correctness label for each claim. 
To localize drift, we define \emph{first-error depth} as the depth (in $G$) of the earliest incorrect claim.

\paragraph{Flow-based Reasoning Metrics.}
We aim to quantify the structure of the \emph{logical flow} induced by the LRM's reasoning. To do so, we model its reasoning process as an \emph{absorbing Markov chain} \cite{kemeny1960finite} on a directed acyclic graph (DAG), where claims extracted from the reasoning trace are transient states. Let $d^-(v)$ and $d^+(v)$ denote the in- and out-degrees of a node $v\in V$, respectively, and let $G_\text{abs}=(V\cup \{a\}, E \cup \{(v,a) \mid v\in V, \, d^+(v) = 0\})$
denote the augmented graph obtained by connecting every leaf node in $G$ to a single absorbing state $a\not\in V$, into which all logical flow converges.
The structure of the flow is characterized by the transition matrix $P$, i.e.,  the row-normalized adjacency matrix of $G_\text{abs}$, whose canonical form can be partitioned as follows (assuming $a$ comes after $v_{|V|}$ in the topological order):
$$P = \begin{pmatrix} Q\in\mathbb{R}^{|V| \times |V|} & R\in\mathbb{R}^{|V| \times 1}  \\ \mathbf{0}\in \mathbb{R}^{1 \times |V|} & 1 \end{pmatrix},$$
with $Q$ and $R$ the transient and absorbing components of the Markov chain, respectively. 
We model the initial distribution of the LRM's \emph{logical mass} on a puzzle instance as a uniform distribution over the source nodes $V_{\textsc{src}}=\{v\in V \mid d^-(v)=0\}$. Formally, we define the initial logical mass as a row vector $\boldsymbol{\pi}$ lying on the $|V|$-dimensional simplex, where $\pi(v)= \nicefrac{1}{|V_{\textsc{src}}|}$ for $v\in V_{\textsc{src}}$ and $\pi(v)=0$ for $v\in V \setminus V_{\textsc{src}}$. Under standard Markov chain dynamics, the distribution $\boldsymbol{\pi}$ evolves through the transient nodes according to $\boldsymbol{\pi}Q^t$, with $t\geq 0$ the discrete time step. The amount of logical mass flowing through each transient node can therefore be quantified as $\mathbf{m} = \boldsymbol{\pi} N,$ with $N=\sum_{t=0}^\infty Q^t$ the fundamental matrix of $G_\text{abs}$. Taking inspiration from structural information theory \cite{li2016structural, shannon1948mathematical}, we quantify the structural entropy of the LRM's logical flow as 
\begin{equation}
    H_{\text{str}}(G) = -\sum_{v\in V} \frac{m(v)}{\Vert\mathbf{m}\Vert} \log \left( \frac{m(v)}{\Vert\mathbf{m}\Vert} \right),
\end{equation}
with $\Vert \mathbf{m}\Vert = \sum_{v \in V} m(v)$. The intuition behind $H_{\text{str}}$ lies in its ability to measure the dispersion versus focus of the LRM's reasoning process across the logical graph. By treating the normalized logical mass as a probability distribution, the entropy quantifies whether the model’s attention is concentrated on a single, deterministic path (low entropy) or scattered across many competing claims and inferences (high entropy). It also penalizes restated and unused claims, since both diffuse the logical flow.

\paragraph{Reasoning Flow Efficiency.}
 We aim to evaluate the LRM's ability to minimize ``structural noise" during its reasoning. To that end, given a puzzle instance, its solution $C^*$ and the LRM reasoning graph $G$, we define the \emph{reasoning flow efficiency}
 \begin{equation}
 \label{eq:rfe}
     \eta = \frac{\log|V| - H_{\text{str}}(G)}{\log|V| - \log|C^*|}.
 \end{equation}
 Intuitively, $\eta$ measures how concentrated the model's logical flow is relative to the minimal claim set that specifies the solution. For solved instances, $\eta$ is designed to lie in $[0,1]$, with $\eta \approx 1$ for highly focused reasoning (low structural entropy relative to the gap between $\log|V|$ and $\log|C^*|$) and smaller values for more diffuse reasoning that spreads flow across many peripheral claims.

\subsection{Reasoning Graph Construction}
\label{sec:graph_construction}

We now describe how reasoning graphs are extracted from model-generated textual traces. The procedure is fully automated, puzzle-agnostic at the graph level, and modular with respect to claim definitions, inference rules, and verification logic. The construction proceeds in three stages: claim extraction, rule extraction, and claim verification (see \cref{fig:graph_extraction_process} for an overview). The extraction pipeline is designed to maximize claim recall while conservatively attributing inference structure, ensuring that all extracted elements are grounded in the text.

\paragraph{Claim Extraction.}
\label{sec:claim_extraction}

We extract claims using a two-pass hybrid procedure that combines deterministic pattern matching with LLM-based extraction. This design balances precision and recall while reducing the reliance on a single extractor.
The reasoning trace is first segmented into sentences and processed in token-balanced chunks to satisfy context-length constraints. For each chunk, we generate candidate claims using two complementary mechanisms:  
(i) a high-precision deterministic extractor followed by an LLM-based completion step that repairs schema violations and adds directly implied missing fields, and  
(ii) a high-recall LLM-based extractor operating without rules.
All extracted claims are treated as event-level occurrences and retain references to their source sentences. Exact-duplicate candidates are removed locally, after which claims are verified in batches of 200 by an LLM with access to a localized support window around their source text. Unsupported or ill-formed claims are discarded, and conservative normalization is applied without collapsing repeated events. Finally, claims are globally deduplicated, ordered by trace position, and assigned identifiers.
The prompts used to extract the claims are included in \cref{app:claim_extraction}, while the puzzle-specific claim types are detailed in \cref{app:claims_rules}.

\paragraph{Rule Extraction.}
\label{sec:rule_extraction}

For each finalized, non-tentative claim, we attempt to recover a single explicit rule application that explains how the claim follows from earlier claims in the trace. Rule extraction is performed once per distinct claim content to avoid redundant derivations. Claims are processed in trace order. For each claim, we construct a truncated prefix of the reasoning trace ending at its last supporting sentence, together with all previously extracted and formatted claims. 
Given this context, an LLM is then prompted to either (i) return a single rule application linking the claim to prior claims, or (ii) return no rule, in which case the claim is treated as directly stated. When a derivation is identified but required premises are missing from the trace, an explicit placeholder claim is inserted to mark the gap. If the same claim has already been stated one, it is classified as restated and a direct edge link the last occurrence of that claim to the restated one.
The prompts used to extract the rules are provided in \cref{app:rule_extraction}, while the puzzle-specific rule types are detailed in \cref{app:claims_rules}.

\paragraph{Claim Verification.}
\label{sec:claim_verification}

Each extracted verifiable claim is independently verified against the executable puzzle environment. For each claim, the environment deterministically checks whether the claim is consistent with the puzzle rules, constraints and final solution. Verification results are attached to claim nodes. See \cref{app:verification} for implementation details.

\section{Experiments}
\label{sec:experiments}

\paragraph{Overview.}
We evaluate large reasoning models on a scalable suite of deterministic grid puzzles and report both outcome metrics and structural metrics extracted from the intermediate traces.
Our main benchmark results are summarized in \cref{tab:benchmark_accuracy}, which reports accuracy and token usage across difficulty levels.
We additionally visualize efficiency scaling trends in \cref{fig:efficiency_scaling}, illustrate qualitative differences between diffuse and focused reasoning graphs in \cref{fig:graph_examples}, and summarize correlations between graph-based efficiency and other trace statistics in \cref{fig:efficiency_correlations}.

\subsection{Benchmark}
\label{sec:benchmark}

We evaluate models on a fixed suite of 21 puzzles (overview in \cref{fig:puzzle_grid}) with four difficulty levels (\emph{Trivial}, \emph{Human easy}, \emph{Human normal}, \emph{Human hard}).
Difficulty is controlled by puzzle-specific generators and corresponds to grid sizes and clue densities that vary by puzzle family (see~\cref{app:puzzle_descriptions}).
For each puzzle and difficulty level, we evaluate on a fixed set of 5 instances and reuse the same instance IDs across all models to support fair comparisons.
All models are prompted with the same solver prompt (see \cref{app:solver_prompt}) and must output a final solution in a standardized format.
We decode using temperature sampling with $T=1$. Because our analysis targets the reasoning \emph{structure} rather than the peak capability, the framework is decoding-agnostic. We fix $T=1$ for consistency, noting that very low temperatures can induce repetitive loops in reasoning models \cite{pipis2025loop} and that decoding strategy can shift model rankings \cite{song2025nondeterminism}.
Outputs are parsed into structured actions/placements. Malformed outputs are treated as incorrect, and final solutions are verified by the executable puzzle environment (\cref{app:verification}). We also report completion token usage as Tokens. 
We evaluate frontier and open reasoning models, including GPT-5 \cite{openai2025gpt5} and open models such as Qwen 3 235B \cite{yang2025qwen3}, DeepSeek V3.2 \cite{deepseekai2025deepseekv32}, and Kimi K2 \cite{kimiteam2025kimik2}.
We report accuracy and token usage by difficulty in \cref{tab:benchmark_accuracy}.
Accuracy is computed over all runs.
To complement the table, \cref{fig:efficiency_scaling} summarizes how efficiency changes with puzzle size across models.
We provide per-puzzle breakdowns in \cref{app:detailed_performance}.

\subsection{Reasoning Graph Statistics}
\label{sec:experiments_graph_stats}
We analyze the structure of intermediate traces by extracting claim graphs as described in \cref{sec:graph_construction}. We use GPT-5.2 for claim extraction and screening, and GPT-5-mini for rule extraction \cite{openai2025gpt5}. We extract graphs only from open-source reasoning models, since closed-source models usually do not provide access to their traces. We report a stability analysis of the extraction pipeline in \cref{sec:graph-extraction-stability}.

\subsection{Reasoning Efficiency}
\label{sec:experiments_efficiency}

\paragraph{Efficiency from claim graphs.}
We compute reasoning-flow efficiency $\eta$ from the extracted claim graphs on solved puzzle instances.
Intuitively, higher $\eta$ indicates more concentrated logical flow relative to the minimal claim set, while lower $\eta$ indicates more diffuse exploration and verification overhead.
\cref{fig:efficiency_correlations} summarizes how $\eta$ relates to verbosity (Tok.), claim composition, claim-level correctness, the depth at which the first incorrect claim occurs, and verification overhead.

\begin{figure}[t]
\centering
\includegraphics[width=0.9\columnwidth]{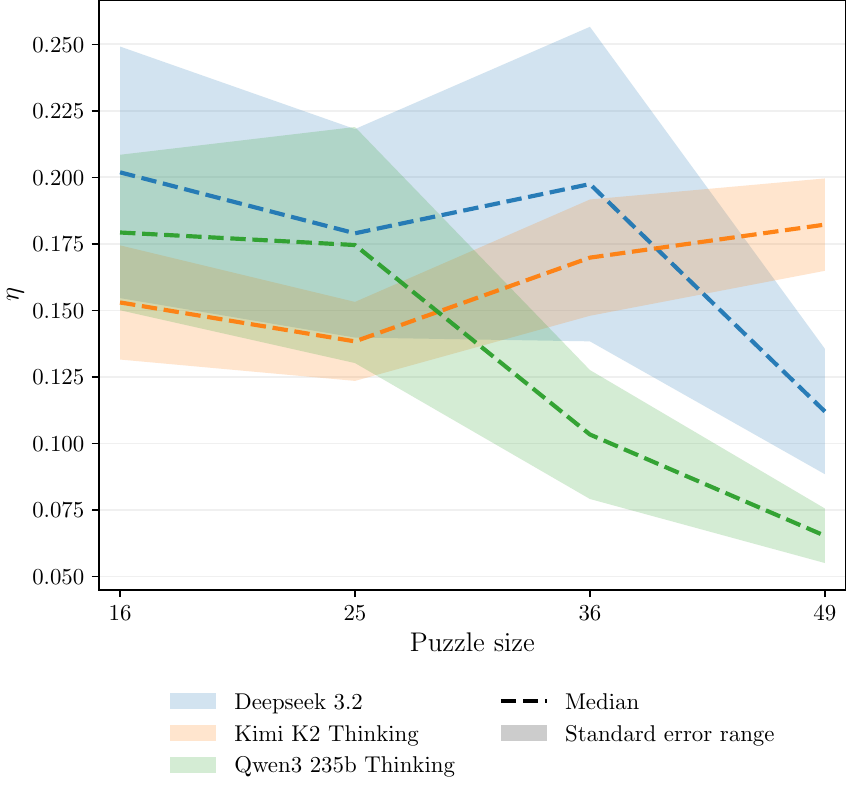}
\caption{\textbf{Reasoning-flow efficiency vs.\ puzzle size under perfect accuracy.}
Mean reasoning-flow efficiency $\eta$ as a function of puzzle size for settings where all puzzle sizes are
solved. We observe that even when correctness is
saturated, $\eta$ exposes differences in reasoning structure and scaling behavior, demonstrating that
accuracy alone hides substantial variation in how solutions are produced.}

\label{fig:efficiency_scaling}
\end{figure}

\section{Discussion}

\paragraph{Difficulty scaling exposes steep accuracy drop despite increased compute.}
Across all models, accuracy declines sharply as difficulty increases from \textit{Trivial} to
\textit{Human hard} (e.g., GPT-5 drops from $83.8\%$ to $5.7\%$; Qwen3~235B drops from $69.5\%$ to $0\%$;
DeepSeek~V3.2 drops from $76.2\%$ to $0\%$; Kimi~K2 drops from $77.1\%$ to $0.95\%$), while mean completion
tokens rise substantially (from roughly $4$-$11$k to approximately $20$-$61$k). Notably, more tokens do
not imply better performance. Kimi~K2 consistently uses the largest token budgets yet does not outperform
GPT-5, whereas GPT-5 achieves the best accuracy at every difficulty while remaining the most
token-efficient. DeepSeek~V3.2 is competitive at \textit{Human normal} ($51.4\%$) but collapses at
\textit{Human hard} despite increased tokens, and Qwen3~235B degrades earlier (only $21.0\%$ at
\textit{Human normal}) and reaches $0\%$ on \textit{Human hard}. Overall, the hardest regime remains
largely unsolved for all models even with large token budgets, suggesting fundamental limitations in
scaling reasoning beyond simply allocating more computation.

\paragraph{Token count is not a proxy for reasoning quality.}
We find that token count alone is a poor proxy for reasoning quality. Across runs, reasoning-flow efficiency $\eta$
is essentially uncorrelated with token count ($r=-0.05, p=0.64$). This suggests that verbosity
cannot be interpreted as either better or worse reasoning structure. Bootstrap $95\%$ confidence intervals for all $\eta$ correlations are reported in \cref{app:structural_details}.

\paragraph{$\eta$ versus simpler structural metrics.}
We test whether $\eta$ adds anything beyond elementary graph statistics
(\cref{tab:simpler_metrics}). Width, $\lvert V \rvert$,
and token count all correlate \emph{negatively} with accuracy and strongly with
one another, indicating that they primarily track puzzle difficulty: harder
instances yield larger, wider graphs with more tokens. $\eta$ is the only metric
that correlates \emph{positively} with accuracy while remaining uncorrelated
with token count, because its size normalization (\cref{eq:rfe}) factors out
graph scale. Per-model breakdowns are consistent (\cref{tab:simpler_metrics_permodel}).

\begin{table}[t]
\centering\small
\caption{\textbf{$\eta$ versus simpler structural metrics.} Pooled Pearson
correlations with claim accuracy and with $\eta$, computed on the same graphs.}
\label{tab:simpler_metrics}
\begin{tabularx}{\columnwidth}{l YY}
\toprule
Metric & vs.\ Accuracy & vs.\ $\eta$ \\
\midrule
Depth            & $-0.263$ & $+0.046$ \\
Diameter         & $-0.329$ & $+0.010$ \\
Avg.\ path length& $-0.182$ & $+0.051$ \\
Width            & $-0.618$ & $-0.431$ \\
$\lvert V \rvert$ & $-0.666$ & $-0.419$ \\
Tokens           & $-0.576$ & $-0.120$ \\
$\eta$           & $+0.368$ & --- \\
\bottomrule
\end{tabularx}
\end{table}

\paragraph{Extra tokens mainly translate into verification overhead.}
Longer traces do not reliably improve $\eta$. These longer traces primarily manifest as
increased verification overhead. In particular, token count correlates strongly with the ratio
$|V_{\mathrm{ver}}|/|V_{\mathrm{sol}}|$ ($r=0.53, p=3\cdot 10^{-9}$), indicating that
additional tokens tend to be spent on checking and auxiliary reasoning around the solution rather
than on expanding the core solution-supporting chain.

\paragraph{Efficiency reflects on-solution focus and penalizes graph bloat.}
In contrast, $\eta$ is closely linked to the amount the reasoning graph contributes to the
solution. Efficiency increases considerably with the solution-supporting portion of the graph ($r=0.55, p=1.1\cdot 10^{-9}$), while it decreases as the overall graph grows ($r=-0.33, p=0.001$). This pattern is consistent with ``graph bloat'' from branching, detours, and
redundant structure from diffuse reasoning flow.

\paragraph{Correctness helps, but structure matters beyond accuracy.}
Correctness is beneficial, but not sufficient. Claim accuracy shows only a moderate association with
$\eta$ ($r=0.33, p=8.9\cdot 10^{-4}$), suggesting that models can be largely correct but
still inefficient if they accumulate many correct claims that are not necessary. 

\paragraph{Early errors are associated with inefficient traces.}
Reasoning traces where the first incorrect claim appears later tend to be more efficient ($r=0.28,
p=0.015$), aligning with the intuition that early errors induce drift and trigger corrective
exploration that lowers reasoning efficiency.

\paragraph{Some redundancy may be beneficial.}
We find that not all redundancy is detrimental. We observe a positive correlation between $\eta$ and the average
number of restatements per unique claim ($r=0.27, p=0.0078$). This may reflect beneficial
``anchoring'' behavior in the form of structured recap of key constraints rather than aimless repetition.

\paragraph{Accuracy and on-solution reasoning reinforce each other.}
Claim accuracy correlates strongly with the solution-supporting fraction of the graph ($r=0.58,
p=2\cdot 10^{-11}$), implying that off-solution exploration is a major source of errors. While these
trends hold globally, model-wise fits differ in slope and intercept across several plots, highlighting
that different models allocate tokens differently (solving vs.\ verifying vs.\ meandering). This finding reinforces that raw token counts are not directly comparable as a measure of reasoning quality.

\paragraph{The signal extends below the solved subset.}
While $\eta$ is defined on solved instances, a comparison against failed
\emph{Tents} traces (\cref{app:structural_details}) shows the same
structure-quality relationship beyond final-answer correctness: failed traces are
substantially less efficient ($\eta$ lower by more than half) and spread reasoning
across larger, more diffuse graphs, with the first error emerging earlier. This
indicates that $\eta$ reflects reasoning quality broadly rather than only
separating solved from unsolved cases.

\section{Limitations}
\label{sec:limitations}
A limitation of our approach is that claim and graph extraction uses LLM components and is therefore not fully static. We reduce bias by separating roles: GPT-5.2 performs semantically difficult claim extraction, while GPT-5-mini handles model-agnostic rule extraction independent of the evaluated systems. As shown in \cref{sec:graph-extraction-stability}, variance is modest and traces with little explicit reasoning yield no meaningful graph, a failure mode that our metrics explicitly capture.

We validate this pipeline empirically: a six-extractor ablation (\cref{sec:graph-extraction-stability}) yields consistent directional trends with no self-bias ($\eta$ varies by only 1.9\% across extractors), manual inspection of 200 rule applications finds 75.5\% fully correct under a strict criterion (a single missing premise counts as a full error), and $\eta$ is robust to graph perturbations (single-perturbation CV below 5\% for $6\times6$ and larger graphs). A second limitation is that each puzzle family requires domain-specific claim and rule types. This is inherent to any process-level evaluation of reasoning traces. Process reward models likewise need per-task verifiers, whereas our structural layer (graph construction, Markov chain, $\eta$) is fully puzzle-agnostic.

\section{Conclusion}
\label{sec:conclusion}
This work proposes a scalable benchmark of deterministic 2D grid puzzles together with a pipeline that converts free-form reasoning traces into verifiable reasoning graphs of atomic claims and dependencies. Our proposed efficiency metric $\eta$ captures how concentrated a model's logical flow is relative to the minimal claim set defining a solution. We demonstrate this framework on \emph{Tents}, where graph structure and $\eta$ distinguish focused deductions from diffuse exploration that accuracy and token count cannot easily capture.

Looking ahead, the structural layer (graph construction, the Markov chain, and $\eta$) is domain-agnostic; only the claim verification is puzzle-specific. The framework therefore extends to any setting where intermediate steps can be checked: mathematical reasoning, where intermediate equations admit symbolic verification, and code generation, where unit tests provide partial claim verification. Agentic tool-use settings, in which inefficient exploration is common, are a further natural target. If extraction becomes sufficiently reliable and low-latency, graph-based signals such as $\eta$ could also serve as auxiliary feedback in post-training with verifiable environments, e.g., a shaping reward favoring solution-focused reasoning while preserving correctness.

\section*{Impact Statement}
This work enables more diagnostic evaluation of LLM reasoning by converting free-form puzzle traces into verifiable reasoning graphs grounded in an executable environment and introducing a structure-sensitive metric that distinguishes solving from meandering or excessive verification beyond accuracy and token count. It can improve reproducibility, help identify early-error cascades and inefficient behaviors, and guide development toward more reliable and cost-effective reasoning. Risks include metric gaming, overfitting research to puzzle-style reasoning, potential bias from imperfect LLM-based extraction, and added compute/latency costs. These are mitigated by treating metrics as diagnostic (not a single leaderboard score) and reporting multiple additional metrics.

\bibliography{references}

@book{kemeny1960finite,
  author    = {Kemeny, J. G. and Snell, J. L.},
  title     = {{Finite Markov Chains}},
  year      = {1960},
  publisher = {D. Van Nostrand Company}
}

@article{li2016structural,
  title   = {{Structural Information and Dynamical Complexity of Networks}},
  author  = {Li, A. and Pan, Y.},
  journal = {{IEEE Transactions on Information Theory}},
  volume  = {62},
  number  = {6},
  pages   = {3290--3339},
  year    = {2016}
}

@article{shannon1948mathematical,
  title   = {{A Mathematical Theory of Communication}},
  author  = {Shannon, C. E.},
  journal = {{The Bell System Technical Journal}},
  volume  = {27},
  number  = {3},
  pages   = {379--423},
  year    = {1948}
}

@inproceedings{chen2025do,
  title     = {{Do NOT Think That Much for 2+3=? On the Overthinking of Long Reasoning Models}},
  author    = {Chen, X. and Xu, J. and Liang, T. and He, Z. and Pang, J. and Yu, D. and others},
  booktitle = {{International Conference on Machine Learning (ICML)}},
  year      = {2025}
}

@inproceedings{wang2025thoughts,
  title     = {{Thoughts Are All Over the Place: On the Underthinking of Long Reasoning Models}},
  author    = {Wang, Y. and Liu, Q. and Xu, J. and Liang, T. and Chen, X. and He, Z. and others},
  booktitle = {{Advances in Neural Information Processing Systems (NeurIPS)}},
  year      = {2025}
}

@inproceedings{dang2025assessing,
  title     = {{Assessing Diversity Collapse in Reasoning}},
  author    = {Dang, X. and Baek, C. and Kolter, J. Z. and Raghunathan, A.},
  booktitle = {{ICLR 2025 Workshop on Scaling Self-Improving Foundation Models without Human Supervision (SSI-FM)}},
  year      = {2025}
}

@inproceedings{fan2025missing,
  title     = {{Missing Premise exacerbates Overthinking: Are Reasoning Models losing Critical Thinking Skill?}},
  author    = {Fan, C. and Li, M. and Sun, L. and Zhou, T.},
  booktitle = {{Conference on Language Modeling (COLM)}},
  year      = {2025}
}

@inproceedings{zhou2025landscape,
  title     = {{Landscape of Thoughts: Visualizing the Reasoning Process of Large Language Models}},
  author    = {Zhou, Z. and Zhu, Z. and Li, X. and Galkin, M. and Feng, X. and Koyejo, S. and others},
  booktitle = {{International Conference on Learning Representations (ICLR)}},
  year      = {2026}
}

@inproceedings{xiong2025mapping,
  title     = {{Mapping the Minds of LLMs: A Graph-Based Analysis of Reasoning LLMs}},
  author    = {Xiong, Z. and Cai, Y. and Li, Z. and Wang, Y.},
  booktitle = {{Conference on Empirical Methods in Natural Language Processing (EMNLP)}},
  year      = {2025}
}

@inproceedings{minegishi2025topology,
  title     = {{Topology of Reasoning: Understanding Large Reasoning Models through Reasoning Graph Properties}},
  author    = {Minegishi, G. and Furuta, H. and Kojima, T. and Iwasawa, Y. and Matsuo, Y.},
  booktitle = {{Advances in Neural Information Processing Systems (NeurIPS)}},
  year      = {2025}
}

@misc{feng2025what,
  title         = {{What Characterizes Effective Reasoning? Revisiting Length, Review, and Structure of CoT}},
  author        = {Feng, Y. and Kempe, J. and Zhang, C. and Jain, P. and Hartshorn, A.},
  year          = {2025},
  eprint        = {2509.19284},
  archivePrefix = {arXiv},
  note          = {arXiv:2509.19284}
}

@inproceedings{zhang2025dag,
  title     = {{DAG-Math: Graph-of-Thought Guided Mathematical Reasoning in LLMs}},
  author    = {Zhang, Y. and Kuzborskij, I. and Lee, J. D. and Leng, C. and Liu, F.},
  booktitle = {{International Conference on Learning Representations (ICLR)}},
  year      = {2026}
}

@misc{zeng2025rejump,
  title         = {{ReJump: A Tree-Jump Representation for Analyzing and Improving LLM Reasoning}},
  author        = {Zeng, Y. and Zhang, S. and Kang, W. and Wu, S. and Zou, L. and Fan, Y. and others},
  year          = {2025},
  eprint        = {2512.00831},
  archivePrefix = {arXiv},
  note          = {arXiv:2512.00831}
}

@inproceedings{wei2022chain,
  title     = {{Chain-of-Thought Prompting Elicits Reasoning in Large Language Models}},
  author    = {Wei, J. and Wang, X. and Schuurmans, D. and Bosma, M. and Ichter, B. and Xia, F. and others},
  booktitle = {{Advances in Neural Information Processing Systems (NeurIPS)}},
  year      = {2022}
}

@inproceedings{yao2023tree,
  title     = {{Tree of Thoughts: Deliberate Problem Solving with Large Language Models}},
  author    = {Yao, S. and Yu, D. and Zhao, J. and Shafran, I. and Griffiths, T. and Cao, Y. and others},
  booktitle = {{Advances in Neural Information Processing Systems (NeurIPS)}},
  year      = {2023}
}

@inproceedings{besta2024graph,
  title     = {{Graph of Thoughts: Solving Elaborate Problems with Large Language Models}},
  author    = {Besta, M. and Blach, N. and Kubicek, A. and Gerstenberger, R. and Podstawski, M. and Gianinazzi, L. and others},
  booktitle = {{AAAI Conference on Artificial Intelligence (AAAI)}},
  year      = {2024}
}

@misc{pandey2025adaptive,
  title         = {{Adaptive Graph of Thoughts: Test-Time Adaptive Reasoning Unifying Chain, Tree, and Graph Structures}},
  author        = {Pandey, T. and Ghukasyan, A. and Goktas, O. and Radha, S. K.},
  year          = {2025},
  eprint        = {2502.05078},
  archivePrefix = {arXiv},
  note          = {arXiv:2502.05078}
}

@misc{openai2024o1,
  title         = {{OpenAI o1 System Card}},
  author        = {OpenAI},
  year          = {2024},
  eprint        = {2412.16720},
  archivePrefix = {arXiv},
  note          = {arXiv:2412.16720}
}

@article{deepseek2025r1,
  title   = {{DeepSeek-R1 Incentivizes Reasoning in LLMs Through Reinforcement Learning}},
  author  = {DeepSeek-AI},
  journal = {{Nature}},
  volume  = {645},
  pages   = {633--638},
  year    = {2025}
}

@misc{chen2021evaluating,
  title         = {{Evaluating Large Language Models Trained on Code}},
  author        = {Chen, M. and Tworek, J. and Jun, H. and Yuan, Q. and Pinto, H. and Kaplan, J. and others},
  year          = {2021},
  eprint        = {2107.03374},
  archivePrefix = {arXiv},
  note          = {arXiv:2107.03374}
}

@misc{lambert2024tulu3,
  title         = {{Tulu 3: Pushing Frontiers in Open Language Model Post-Training}},
  author        = {Lambert, N. and Morrison, J. and Pyatkin, V. and Huang, S. and Ivison, H. and Brahman, F. and others},
  year          = {2024},
  eprint        = {2411.15124},
  archivePrefix = {arXiv},
  note          = {arXiv:2411.15124}
}

@misc{amodei2016concrete,
  title         = {{Concrete Problems in AI Safety}},
  author        = {Amodei, D. and Olah, C. and Steinhardt, J. and Christiano, P. and Schulman, J. and Man\'{e}, D.},
  year          = {2016},
  eprint        = {1606.06565},
  archivePrefix = {arXiv},
  note          = {arXiv:1606.06565}
}

@misc{yang2025qwen3,
  title         = {{Qwen3 Technical Report}},
  author        = {Yang, A. and others},
  year          = {2025},
  eprint        = {2505.09388},
  archivePrefix = {arXiv},
  note          = {arXiv:2505.09388}
}

@misc{deepseekai2025deepseekv32,
  title         = {{DeepSeek-V3.2: Pushing the Frontier of Open Large Language Models}},
  author        = {DeepSeek-AI and others},
  year          = {2025},
  eprint        = {2512.02556},
  archivePrefix = {arXiv},
  note          = {arXiv:2512.02556}
}

@misc{kimiteam2025kimik2,
  title         = {{Kimi K2: Open Agentic Intelligence}},
  author        = {Kimi Team and others},
  year          = {2025},
  eprint        = {2507.20534},
  archivePrefix = {arXiv},
  note          = {arXiv:2507.20534}
}

@misc{openai2025gpt5,
  title         = {{OpenAI GPT-5 System Card}},
  author        = {OpenAI},
  year          = {2025},
  eprint        = {2601.03267},
  archivePrefix = {arXiv},
  note          = {arXiv:2601.03267}
}

@inproceedings{mirzadeh2025gsm,
  title     = {{GSM-Symbolic: Understanding the Limitations of Mathematical Reasoning in Large Language Models}},
  author    = {Mirzadeh, I. and Alizadeh, K. and Shahrokhi, H. and Tuzel, O. and Bengio, S. and Farajtabar, M.},
  booktitle = {{International Conference on Learning Representations (ICLR)}},
  year      = {2025}
}

@inproceedings{lin2025zebralogic,
  title     = {{ZebraLogic: On the Scaling Limits of LLMs for Logical Reasoning}},
  author    = {Lin, B. Y. and Le Bras, R. and Richardson, K. and Sabharwal, A. and Poovendran, R. and Clark, P. and others},
  booktitle = {{International Conference on Machine Learning (ICML)}},
  year      = {2025}
}

@inproceedings{estermann2024puzzles,
  title     = {{PUZZLES: A Benchmark for Neural Algorithmic Reasoning}},
  author    = {Estermann, B. and Lanzend\"{o}rfer, L. and Niedermayr, Y. and Wattenhofer, R.},
  booktitle = {{Advances in Neural Information Processing Systems (NeurIPS)}},
  year      = {2024}
}

@inproceedings{berdoz2026text,
  title     = {{Text-to-Scene with Large Reasoning Models}},
  author    = {Berdoz, F. and Lanzend\"{o}rfer, L. and Tuninga, N. and Wattenhofer, R.},
  booktitle = {{AAAI Conference on Artificial Intelligence (AAAI)}},
  year      = {2026}
}

@inproceedings{tafjord2021proofwriter,
  title     = {{ProofWriter: Generating Implications, Proofs, and Abductive Statements over Natural Language}},
  author    = {Tafjord, O. and Dalvi, B. and Clark, P.},
  booktitle = {{Findings of the Association for Computational Linguistics: ACL-IJCNLP}},
  year      = {2021}
}

@inproceedings{saparov2023language,
  title     = {{Language Models Are Greedy Reasoners: A Systematic Formal Analysis of Chain-of-Thought}},
  author    = {Saparov, A. and He, H.},
  booktitle = {{International Conference on Learning Representations (ICLR)}},
  year      = {2023}
}

@inproceedings{parmar2024logicbench,
  title     = {{LogicBench: Towards Systematic Evaluation of Logical Reasoning Ability of Large Language Models}},
  author    = {Parmar, M. and Patel, N. and Varshney, N. and Nakamura, M. and Luo, M. and Mashetty, S. and others},
  booktitle = {{Annual Meeting of the Association for Computational Linguistics (ACL)}},
  year      = {2024}
}

@misc{cobbe2021training,
  title         = {{Training Verifiers to Solve Math Word Problems}},
  author        = {Cobbe, K. and Kosaraju, V. and Bavarian, M. and Chen, M. and Jun, H. and Kaiser, L. and others},
  year          = {2021},
  eprint        = {2110.14168},
  archivePrefix = {arXiv},
  note          = {arXiv:2110.14168}
}

@inproceedings{wu2024smartplay,
  title     = {{SmartPlay: A Benchmark for LLMs as Intelligent Agents}},
  author    = {Wu, Y. and Tang, X. and Mitchell, T. and Li, Y.},
  booktitle = {{International Conference on Learning Representations (ICLR)}},
  year      = {2024}
}

@misc{zhang2025puzzlebench,
  title         = {{PuzzleBench: A Fully Dynamic Evaluation Framework for Large Multimodal Models on Puzzle Solving}},
  author        = {Zhang, Z. and Chen, Z. and Zhang, Z. and Sun, Y. and Tian, Y. and Jia, Z. and others},
  year          = {2025},
  eprint        = {2504.10885},
  archivePrefix = {arXiv},
  note          = {arXiv:2504.10885}
}

@inproceedings{chen2025enigmata,
  title     = {{Enigmata: Scaling Logical Reasoning in Large Language Models with Synthetic Verifiable Puzzles}},
  author    = {Chen, J. and He, Q. and Yuan, S. and Chen, A. and Cai, Z. and Dai, W. and others},
  booktitle = {{Advances in Neural Information Processing Systems (NeurIPS)}},
  year      = {2025}
}

@inproceedings{shojaee2025illusion,
  title     = {{The Illusion of Thinking: Understanding the Strengths and Limitations of Reasoning Models via the Lens of Problem Complexity}},
  author    = {Shojaee, P. and Mirzadeh, I. and Alizadeh, K. and Horton, M. and Bengio, S. and Farajtabar, M.},
  booktitle = {{Advances in Neural Information Processing Systems (NeurIPS)}},
  year      = {2025}
}

@inproceedings{wei2025satbench,
  title     = {{SATBench: Benchmarking LLMs' Logical Reasoning via Automated Puzzle Generation from SAT Formulas}},
  author    = {Wei, A. and Wu, Y. and Wan, Y. and Suresh, T. and Tan, H. and Zhou, Z. and others},
  booktitle = {{Conference on Empirical Methods in Natural Language Processing (EMNLP)}},
  year      = {2025}
}

@misc{tatham2025portable,
  title  = {{Simon Tatham's Portable Puzzle Collection}},
  author = {Tatham, S.},
  year   = {2025},
  url    = {https://www.chiark.greenend.org.uk/~sgtatham/puzzles/},
  note   = {Accessed January 28, 2026}
}

@article{yato2003complexity,
  title   = {{Complexity and Completeness of Finding Another Solution and Its Application to Puzzles}},
  author  = {Yato, T. and Seta, T.},
  journal = {{IEICE Transactions on Fundamentals of Electronics, Communications and Computer Sciences}},
  volume  = {E86-A},
  number  = {5},
  pages   = {1052--1060},
  year    = {2003}
}

@misc{debiasi2012complexity,
  title  = {{The Complexity of Camping}},
  author = {De Biasi, M.},
  year   = {2012},
  url    = {https://www.nearly42.org/vdisk/cstheory/tentsnpc2.pdf},
  note   = {Technical report. Accessed May 28, 2026}
}

@misc{mcphail2005light,
  title  = {{Light Up is NP-complete}},
  author = {McPhail, B.},
  year   = {2005},
  url    = {http://mountainvistasoft.com/docs/lightup-is-np-complete.pdf},
  note   = {Technical report. Accessed May 28, 2026}
}

@mastersthesis{dejong2023mosaic,
  title  = {{Mosaic as a SAT Problem}},
  author = {de Jong, T.},
  year   = {2023},
  school = {{Radboud University}},
  type   = {{Bachelor's thesis}}
}

@techreport{ueda1996nonogram,
  title       = {{NP-completeness Results for NONOGRAM via Parsimonious Reductions}},
  author      = {Ueda, N. and Nagao, T.},
  year        = {1996},
  institution = {{Tokyo Institute of Technology}},
  number      = {TR96-0008},
  type        = {{Technical report}}
}

@misc{friedman2002spiral,
  title  = {{Spiral Galaxies Puzzles are NP-complete}},
  author = {Friedman, E.},
  year   = {2002},
  url    = {https://erich-friedman.github.io/papers/spiral.pdf},
  note   = {Unpublished manuscript. Accessed May 28, 2026}
}

@article{kolker2012magnets,
  title   = {{The Magnets Puzzle is NP-Complete}},
  author  = {K\"{o}lker, J.},
  journal = {{Journal of Information Processing}},
  volume  = {20},
  number  = {3},
  pages   = {707--708},
  year    = {2012}
}

@article{kral2004tough,
  title   = {{It Is Tough to Be a Plumber}},
  author  = {Kr\'{a}l, D. and Majerech, V. and Sgall, J. and Tich\'{y}, T. and Woeginger, G.},
  journal = {{Theoretical Computer Science}},
  volume  = {313},
  number  = {3},
  pages   = {474--484},
  year    = {2004}
}

@misc{friedman2002pearl,
  title  = {{Pearl Puzzles are NP-complete}},
  author = {Friedman, E.},
  year   = {2002},
  url    = {https://erich-friedman.github.io/papers/pearl.pdf},
  note   = {Unpublished manuscript. Accessed May 28, 2026}
}

@article{takenaga2013shikaku,
  title   = {{Shikaku and Ripple Effect are NP-Complete}},
  author  = {Takenaga, Y. and Aoyagi, S. and Iwata, S. and Kasai, T.},
  journal = {{Congressus Numerantium}},
  volume  = {216},
  pages   = {119--127},
  year    = {2013}
}

@mastersthesis{maarse2019npcomplete,
  title  = {{The NP-completeness of Some Lesser Known Logic Puzzles}},
  author = {Maarse, M.},
  year   = {2019},
  school = {{Utrecht University}},
  type   = {{Bachelor's thesis}}
}

@misc{debiasi2012binary,
  title  = {{Binary Puzzle is NP-complete}},
  author = {De Biasi, M.},
  year   = {2012},
  url    = {https://www.nearly42.org/vdisk/cstheory/binaryp.pdf},
  note   = {Technical report. Accessed May 28, 2026}
}

@misc{pipis2025loop,
  title  = {{Wait, Wait, Wait... Why Do Reasoning Models Loop?}},
  author = {Pipis, C. and Garg, S. and Kontonis, V. and Shrivastava, V. and Krishnamurthy, A. and Papailiopoulos, D.},
  year   = {2025},
  note   = {arXiv:2512.12895}
}

@inproceedings{song2025nondeterminism,
  title     = {{The Good, The Bad, and The Greedy: Evaluation of LLMs Should Not Ignore Non-Determinism}},
  author    = {Song, Y. and Wang, G. and Li, S. and Lin, B. Y.},
  booktitle = {{Conference of the Nations of the Americas Chapter of the Association for Computational Linguistics (NAACL)}},
  year      = {2025}
}

@misc{muennighoff2025s1,
  title  = {{s1: Simple Test-Time Scaling}},
  author = {Muennighoff, N. and Yang, Z. and Shi, W. and Li, X. L. and Fei-Fei, L. and Hajishirzi, H. and others},
  year   = {2025},
  note   = {arXiv:2501.19393}
}
\bibliographystyle{icml2026}

\newpage
\appendix
\onecolumn
\paragraph{Appendix overview.}
This appendix contains (i) stability checks for the graph-extraction pipeline in \cref{sec:graph-extraction-stability}, (ii) claim verification details in \cref{app:verification}, (iii) per-puzzle performance breakdowns in \cref{app:detailed_performance}, (iv) all prompts used in the pipeline in \cref{app:prompts}, and (v) puzzle rules and difficulty specifications in \cref{app:puzzle_descriptions}.
\section{Graph Extraction Stability}
\label{sec:graph-extraction-stability}

Our pipeline uses LLM-based components, so the extracted graph is not guaranteed to be identical across runs. To quantify extractor variance, we repeat graph extraction three times on the \emph{same} reasoning trace. We report stability of the unique extracted claim set $\mathcal{C}_{\text{unique}}$ (deduplicated by exact payload matching) via the pairwise Jaccard overlap between runs, and we report how this variability propagates to our downstream structural metrics ($H_{\text{str}}$ and $\eta$). Overall, repeated extraction from the same trace yields high claim-set overlap, and $H_{\text{str}}$ and $\eta$ remain comparatively stable (Table~\ref{tab:graph_extraction_stability}), supporting their use as coarse, graph-level descriptors rather than brittle artifacts of extraction.

\begin{table}[h]
\centering
\small
\setlength{\tabcolsep}{3pt}
\renewcommand{\arraystretch}{0.95}
\caption{\textbf{Graph extraction stability (same-trace repeats).} We report stability under repeated extraction from the same reasoning trace (three runs). $\lvert \mathcal{C}_{\text{unique}} \rvert$ is the mean number of unique claims (deduplicated by exact payload matching). Jaccard is the mean $\pm$ standard deviation of pairwise Jaccard overlaps between claim sets. $H_{\text{str}}$ and $\eta$ are reported as mean $\pm$ standard deviation over runs.}\label{tab:graph_extraction_stability}
\begin{tabularx}{0.99\textwidth}{l YYYY}
\toprule
Model
& $\lvert \mathcal{C}_{\text{unique}} \rvert$
& Jaccard
& $H_{\text{str}}$
& $\eta$ \\
\midrule
Kimi-K2 Thinking
& 22.7 & 0.89 $\pm$ 0.08 & 3.184 $\pm$ 0.191 & 0.946 $\pm$ 0.059 \\
DeepSeek V3.2
& 25.0 & 0.79 $\pm$ 0.07 & 3.573 $\pm$ 0.092 & 0.840 $\pm$ 0.022 \\
Qwen3 235B 2507 Thinking
& 37.7 & 0.98 $\pm$ 0.01 & 3.660 $\pm$ 0.063 & 0.820 $\pm$ 0.014 \\
\bottomrule
\end{tabularx}
\end{table}

\subsection{Robustness to Extractor Choice}
\label{app:extractor_ablation}
To verify that our structural findings are not an artifact of using GPT-5.2 as the claim extractor, we repeat extraction with five additional LLMs on the same reasoning traces. Directional trends are consistent across all six extractors: the DeepSeek source yields the lowest $\eta$ for every extractor ($6/6$), and no extractor systematically inflates $\eta$ on traces from its own model family (no self-bias). Across extractors, $\eta$ has a coefficient of variation of $1.9\%$.

\begin{table}[h]
\centering
\small
\setlength{\tabcolsep}{4pt}
\renewcommand{\arraystretch}{0.95}
\caption{\textbf{Robustness to extractor choice.} Graph-extraction statistics
when the claim extractor is varied across six LLMs (rows), applied to traces from
three source models. Values are mean $\pm$ std over three runs.
\emph{Deterministically extracted solution claims are excluded}, so $\eta$ here
isolates model-generated reasoning and is not directly comparable to
\cref{tab:graph_extraction_stability}; Jaccard values are conservative lower
bounds.}
\label{tab:extractor_ablation}
\begin{tabularx}{0.99\textwidth}{l l YYY}
\toprule
Extractor & Source & Jaccard & $H_{\text{str}}$ & $\eta$ \\
\midrule
\multirow{3}{*}{Main (GPT-5.2)}
 & Kimi     & $0.89 \pm 0.08$ & $3.38 \pm 0.09$ & $0.15 \pm 0.06$ \\
 & DeepSeek & $0.79 \pm 0.07$ & $3.39 \pm 0.07$ & $0.14 \pm 0.06$ \\
 & Qwen3    & $0.98 \pm 0.01$ & $3.66 \pm 0.06$ & $0.15 \pm 0.04$ \\
\midrule
\multirow{3}{*}{GPT-5.4}
 & Kimi     & $0.81 \pm 0.07$ & $3.47 \pm 0.25$ & $0.21 \pm 0.08$ \\
 & DeepSeek & $0.98 \pm 0.01$ & $3.40 \pm 0.06$ & $0.13 \pm 0.06$ \\
 & Qwen3    & $0.92 \pm 0.04$ & $3.43 \pm 0.22$ & $0.21 \pm 0.05$ \\
\midrule
\multirow{3}{*}{DeepSeek}
 & Kimi     & $0.88 \pm 0.06$ & $3.45 \pm 0.22$ & $0.20 \pm 0.08$ \\
 & DeepSeek & $0.98 \pm 0.02$ & $3.40 \pm 0.07$ & $0.11 \pm 0.04$ \\
 & Qwen3    & $0.91 \pm 0.01$ & $3.44 \pm 0.20$ & $0.21 \pm 0.05$ \\
\midrule
\multirow{3}{*}{Qwen3}
 & Kimi     & $0.92 \pm 0.03$ & $3.32 \pm 0.03$ & $0.20 \pm 0.10$ \\
 & DeepSeek & $0.90 \pm 0.03$ & $3.41 \pm 0.08$ & $0.10 \pm 0.03$ \\
 & Qwen3    & $0.92 \pm 0.01$ & $3.31 \pm 0.03$ & $0.29 \pm 0.05$ \\
\midrule
\multirow{3}{*}{Kimi}
 & Kimi     & $0.67 \pm 0.06$ & $3.42 \pm 0.22$ & $0.22 \pm 0.10$ \\
 & DeepSeek & $0.90 \pm 0.07$ & $3.37 \pm 0.05$ & $0.12 \pm 0.03$ \\
 & Qwen3    & $0.71 \pm 0.09$ & $3.46 \pm 0.22$ & $0.20 \pm 0.05$ \\
\midrule
\multirow{3}{*}{GPT-OSS}
 & Kimi     & $0.83 \pm 0.07$ & $3.40 \pm 0.21$ & $0.22 \pm 0.09$ \\
 & DeepSeek & $0.65 \pm 0.05$ & $3.38 \pm 0.20$ & $0.21 \pm 0.09$ \\
 & Qwen3    & $0.77 \pm 0.10$ & $3.33 \pm 0.09$ & $0.22 \pm 0.04$ \\
\bottomrule
\end{tabularx}
\end{table}

\subsection{Sensitivity to Markov-Chain Assumptions}
\label{app:markov_sensitivity}
Our flow metric assumes a uniform initial distribution $\boldsymbol{\pi}$ over
source nodes and unweighted (row-normalized) transitions $P$. We test seven
alternatives against this default. All transition-weighting variants preserve
near-perfect instance ranking; initial-distribution variants shift more but
remain strongly rank-correlated. The variation in $\eta$ stems primarily from the
normalization rather than the underlying entropy, confirming that $\eta$ captures
intrinsic graph structure.

\begin{table}[h]
\centering
\small
\setlength{\tabcolsep}{6pt}
\renewcommand{\arraystretch}{0.95}
\caption{\textbf{Sensitivity to Markov-chain modeling choices.} Each row is an alternative to the default (uniform $\boldsymbol{\pi}$, uniform $P$). We report the Spearman rank correlation of per-instance $\eta$ with the default ($\rho$), the mean absolute deviation in $\eta$ ($\lvert\Delta\eta\rvert$), and the mean $\eta$ under the variant.}
\label{tab:markov_sensitivity}
\begin{tabularx}{0.99\textwidth}{l YYY}
\toprule
Variant & $\rho$ & $\lvert\Delta\eta\rvert$ & Mean $\eta$ \\
\midrule
Text-proximity $P$                            & $0.993$ & $0.009$ & $0.149$ \\
Exp-decay $P$                                 & $0.968$ & $0.018$ & $0.145$ \\
Inverse-fan-in $P$                            & $0.964$ & $0.028$ & $0.130$ \\
Subtree-weighted $P$                          & $0.861$ & $0.062$ & $0.218$ \\
Recency $\boldsymbol{\pi}$                    & $0.850$ & $0.118$ & $0.276$ \\
Recency $\boldsymbol{\pi}$ + Text-prox.\ $P$  & $0.847$ & $0.109$ & $0.267$ \\
Degree $\boldsymbol{\pi}$                     & $0.778$ & $0.074$ & $0.228$ \\
\bottomrule
\end{tabularx}
\end{table}

\subsection{Sensitivity to Extraction Granularity}
\label{app:perturbation}
To assess how extraction errors propagate to $\eta$, we apply controlled perturbations to each extracted graph, removing a single node or edge, or adding $1$-$4$ random edges, and report the coefficient of variation of $\eta$ by puzzle size. Sensitivity decreases with graph size; for $6\times6$ graphs and larger, single-perturbation CV is below $5\%$.

\begin{table}[h]
\centering
\small
\setlength{\tabcolsep}{6pt}
\renewcommand{\arraystretch}{0.95}
\caption{\textbf{Sensitivity of $\eta$ to graph perturbations.} Coefficient of
variation of $\eta$ under each perturbation type, by puzzle size.}
\label{tab:perturbation}
\begin{tabularx}{0.99\textwidth}{l YYYY}
\toprule
Perturbation & $4\times4$ & $5\times5$ & $6\times6$ & $7\times7$ \\
\midrule
Node removal   & $0.130$ & $0.079$ & $0.051$ & $0.037$ \\
Edge removal   & $0.115$ & $0.060$ & $0.043$ & $0.035$ \\
Edge add ($1$) & $0.165$ & $0.074$ & $0.051$ & $0.038$ \\
Edge add ($2$) & $0.260$ & $0.107$ & $0.075$ & $0.052$ \\
Edge add ($3$) & $0.300$ & $0.134$ & $0.091$ & $0.071$ \\
Edge add ($4$) & $0.340$ & $0.167$ & $0.118$ & $0.081$ \\
\bottomrule
\end{tabularx}
\end{table}

\section{Environment and Claim Verification}
\label{app:verification}

For each puzzle family, we implement deterministic verifiers that map an extracted claim to a correctness label (correct / incorrect), using the executable environment. Placement claims are checked against the environment's ground-truth solution, while constraint claims are checked against the given puzzle state (clues). State-dependent (non-static) claims are not verified as tracking partial solutions is not trivial and not robust. If a branch is explicitly presented as a contradiction in the trace, we do not count the triggering claim as incorrect (it is used as part of a refutation). Because this logic relies on faithful extraction of commitments and branch boundaries, we use it conservatively to avoid introducing fragility.

\paragraph{Edge validation.}
Whereas claim nodes are verified deterministically against the environment, edges are attributed by constrained LLM rule application and are not programmatically checked. To quantify edge quality, we manually evaluated $200$ randomly sampled rule applications: $151$ ($75.5\%$) were fully correct under a strict criterion in which a single missing premise counts as a full error. Most errors were incomplete derivations (a missing premise) rather than spurious edges.

\paragraph{Unverifiable claims.}
State-dependent (non-static) claims reference the current partial board state and are excluded from verification to avoid cascading grading errors. 
\cref{tab:unverifiable_fraction} reports the fraction of such claims per model and puzzle size; all fractions are below $15\%$. Because errors in these claims typically propagate to later verifiable claims, their practical effect is a slightly delayed first-error detection and a mild attenuation of the $\eta$-accuracy correlation.

\begin{table}[h]
\centering
\small
\setlength{\tabcolsep}{6pt}
\renewcommand{\arraystretch}{0.95}
\caption{\textbf{Fraction of unverifiable (state-dependent) claims.} Mean $\pm$ std by model and puzzle size.}
\label{tab:unverifiable_fraction}
\begin{tabularx}{0.99\textwidth}{l YYYY}
\toprule
Model & $4\times4$ & $5\times5$ & $6\times6$ & $7\times7$ \\
\midrule
Qwen3    & $0.00 \pm 0.00$ & $0.13 \pm 0.16$ & $0.11 \pm 0.11$ & $0.07 \pm 0.08$ \\
DeepSeek & $0.03 \pm 0.06$ & $0.06 \pm 0.10$ & $0.07 \pm 0.07$ & $0.14 \pm 0.14$ \\
Kimi     & $0.00 \pm 0.00$ & $0.13 \pm 0.19$ & $0.08 \pm 0.09$ & $0.14 \pm 0.17$ \\
\bottomrule
\end{tabularx}
\end{table}

\section{Structural Analysis Details}
\label{app:structural_details}
The structural analysis in \cref{fig:efficiency_correlations} and \cref{fig:efficiency_scaling} is computed on $85$ solved \emph{Tents} instances ($3$ traces $\times$ $3$ instances per size per model; $20$/$35$/$30$ for DeepSeek/Kimi/Qwen3), balanced across sizes ($24$/$20$/$21$/$20$ for $4\times4$ through $7\times7$).

\paragraph{Token-quality correlations within model and size.}
To rule out that the near-zero token-$\eta$ relationship is an artifact of pooling across difficulty, we stratify by model and by puzzle size (\cref{tab:token_stratified}). Token-$\eta$ correlations are near zero at every level, while token-accuracy correlations are consistently negative and strengthen with size, consistent with token count proxying for difficulty rather than reasoning quality \cite{muennighoff2025s1, shojaee2025illusion}.

\begin{table}[h]
\centering
\small
\setlength{\tabcolsep}{6pt}
\renewcommand{\arraystretch}{0.95}
\caption{\textbf{Token-quality correlations, stratified.} Pearson correlations of completion tokens with $\eta$ and with accuracy, within each model and within each puzzle size.}
\label{tab:token_stratified}
\begin{tabularx}{0.99\textwidth}{l YY}
\toprule
Stratum & Tokens vs.\ $\eta$ & Tokens vs.\ Acc. \\
\midrule
DeepSeek   & $-0.399$ & $-0.756$ \\
Kimi       & $+0.176$ & $-0.388$ \\
Qwen3      & $-0.397$ & $-0.770$ \\
\midrule
$4\times4$ & $+0.020$ & $-0.324$ \\
$5\times5$ & $-0.028$ & $-0.604$ \\
$6\times6$ & $-0.032$ & $-0.532$ \\
$7\times7$ & $-0.040$ & $-0.771$ \\
\bottomrule
\end{tabularx}
\end{table}

\paragraph{Per-Model Structural-Metric Correlations}
\label{app:simpler_metrics_permodel}
\cref{tab:simpler_metrics} reports correlations pooled across models; here we give
the per-model breakdown. The directional trends are consistent across all three
source models: $\lvert V\rvert$, width, and token count correlate \emph{negatively}
with accuracy for every model, while $\eta$ is the only metric that correlates
\emph{positively} with accuracy throughout. The associations with $\eta$ (right-hand columns) are weaker and vary in sign across models, but every significant $\eta$ correlation is negative and all sign-flipped (positive) values, notably Kimi's entire column, are non-significant.

\begin{table}[h]
\centering\small
\setlength{\tabcolsep}{4pt}
\renewcommand{\arraystretch}{0.95}
\caption{\textbf{Per-model structural-metric correlations.} Pearson correlations of
each structural metric with claim accuracy and with $\eta$, by source model
(cf.\ the pooled values in \cref{tab:simpler_metrics}). Significance:
$^{*}p<0.05$, $^{**}p<0.01$, $^{***}p<0.001$.}
\label{tab:simpler_metrics_permodel}
\begin{tabularx}{0.99\textwidth}{l YY YY YY}
\toprule
& \multicolumn{2}{c}{DeepSeek V3.2}
& \multicolumn{2}{c}{Kimi K2}
& \multicolumn{2}{c}{Qwen3 235B} \\
\cmidrule(lr){2-3}\cmidrule(lr){4-5}\cmidrule(lr){6-7}
Metric & Acc. & $\eta$ & Acc. & $\eta$ & Acc. & $\eta$ \\
\midrule
Depth             & $-0.226$       & $-0.166$     & $-0.403^{*}$  & $+0.272$     & $-0.225$       & $+0.034$ \\
Diameter          & $-0.342$       & $-0.233$     & $-0.394^{*}$  & $+0.158$     & $-0.335$       & $-0.007$ \\
Avg.\ path length & $-0.046$       & $-0.165$     & $-0.296$      & $+0.076$     & $-0.192$       & $+0.156$ \\
Width             & $-0.903^{***}$ & $-0.472^{*}$ & $-0.483^{**}$ & $+0.088$     & $-0.667^{***}$ & $-0.545^{**}$ \\
$\lvert V\rvert$  & $-0.899^{***}$ & $-0.492^{*}$ & $-0.525^{**}$ & $+0.104$     & $-0.708^{***}$ & $-0.551^{**}$ \\
Tokens            & $-0.838^{***}$ & $-0.399$     & $-0.450^{**}$ & $+0.176$     & $-0.770^{***}$ & $-0.397^{*}$ \\
$\eta$            & $+0.440$       & ---          & $+0.174$      & ---          & $+0.456^{*}$   & --- \\
\bottomrule
\end{tabularx}
\end{table}

\paragraph{Efficiency below the solved subset.}
We additionally compare structural metrics on \emph{incorrect} Tents traces against
correct traces at matched puzzle sizes. We exclude failed \emph{generations}
(truncated or aborted runs producing fewer than $\sim\!1$k completion tokens and no
extractable reasoning), which reflect infrastructure failures rather than flawed
reasoning. On the remaining failed traces ($n=10$ incorrect vs.\ $35$ matched
correct, sizes $5\times5$--$7\times7$, three models; \cref{tab:failed_traces}),
reasoning-flow efficiency is $57\%$ lower, claim accuracy is $\sim\!14\%$ lower, and
the first error appears $\sim\!51\%$ earlier. Failed traces are also larger
($+52\%$ nodes), wider ($+73\%$), and higher-entropy, and devote a smaller fraction
of the graph to the solution, indicating that failing models reason more diffusely
from early on rather than making isolated late mistakes. This analysis is limited to
Tents; a broader study across puzzle families is left to future work.

\begin{table}[h]
\centering\small
\setlength{\tabcolsep}{6pt}
\renewcommand{\arraystretch}{0.95}
\caption{\textbf{Correct vs.\ failed traces (Tents).} Pooled means over
matched correct and genuinely failed traces; failed \emph{generations}
($<\!1$k tokens, no extractable reasoning) are excluded. $\eta$ uses the same
(solution-excluded) convention as the main analysis.}
\label{tab:failed_traces}
\begin{tabularx}{0.78\columnwidth}{l YYY}
\toprule
Metric & Correct & Failed & $\Delta$ \\
\midrule
Claim accuracy                 & $0.90$ & $0.78$ & $-14\%$ \\
First-error depth (norm.)      & $0.37$ & $0.18$ & $-51\%$ \\
Efficiency $\eta$              & $0.127$& $0.055$& $-57\%$ \\
Solution fraction $\lvert V_{\mathrm{sol}}\rvert/\lvert V\rvert$ & $0.43$ & $0.33$ & $-23\%$ \\
Graph size $\lvert V\rvert$    & $168$  & $256$  & $+52\%$ \\
Width                          & $113$  & $195$  & $+73\%$ \\
Flow entropy $H_{\text{str}}$  & $4.84$ & $5.24$ & $+8\%$ \\
\bottomrule
\end{tabularx}
\end{table}

\paragraph{Bootstrap confidence intervals.}
For every association involving $\eta$ in \cref{fig:efficiency_correlations}, we compute bootstrap $95\%$ confidence intervals ($5{,}000$ resamples over the $85$ instances above). All intervals exclude zero except the one for token count, consistent with $\eta$ being uncorrelated with verbosity but reliably associated with the other quantities: token count ($r=-0.05$, $[-0.23, 0.13]$), solution-supporting fraction ($r=0.55$, $[0.40, 0.66]$), claim accuracy ($r=0.33$, $[0.16, 0.49]$), first-error depth ($r=0.28$, $[0.03, 0.50]$), restatements per claim ($r=0.27$, $[0.03, 0.50]$), and graph size ($r=-0.33$, $[-0.48, -0.14]$).

\section{Detailed performance}
\label{app:detailed_performance}
This section reports per-puzzle accuracy (solved / total) and mean completion tokens (in parentheses) for each difficulty level, complementing the aggregate benchmark table in the main text (\cref{tab:benchmark_accuracy}). The tables below break results down by model family.

\begin{table}[h]
\centering
\small
\setlength{\tabcolsep}{3pt}
\renewcommand{\arraystretch}{0.9}
\caption{\textbf{Detailed performance (GPT-5).} Per-puzzle results by difficulty on the 21-puzzle suite. Each cell reports \# solved / \# evaluated instances, with mean completion token count in parentheses, computed over the evaluated instances at that difficulty. All runs use the same solver prompt and decoding settings as in the main benchmark (\cref{sec:benchmark}).}
\label{tab:puzzle_details_perf_gpt}

\begin{tabularx}{\textwidth}{YYYYY}
\toprule
\textbf{Puzzle} & \textbf{Trivial} & \textbf{Easy} & \textbf{Normal} & \textbf{Hard}\\
\midrule
Solo        & 5/5 (1270.8)   & 0/5 (798.2)    & 5/5 (7250.4)    & 4/5 (17960.2) \\
Tents       & 5/5 (3165.2)   & 5/5 (7402.8)   & 5/5 (13974.8)   & 0/5 (22825.3) \\
Filling     & 5/5 (3957.6)   & 1/5 (24751.2)  & 0/5 (24745.7)   & 0/5 (16066.4) \\
Lightup     & 5/5 (1657.0)   & 5/5 (7586.8)   & 5/5 (16345.8)   & 0/5 (29066.4) \\
Mosaic      & 5/5 (1205.6)   & 5/5 (3708.0)   & 4/5 (21926.0)   & 0/5 (8904.4) \\
Keen        & 5/5 (2319.2)   & 5/5 (5098.0)   & 5/5 (13839.6)   & 0/5 (40029.6) \\
Range       & 5/5 (3075.0)   & 5/5 (8830.0)   & 4/5 (24656.6)   & 0/5 (24439.8) \\
Pattern     & 5/5 (632.4)    & 5/5 (27481.2)  & 0/5 (25613.6)   & 0/5 (13722.2) \\
Dominosa    & 5/5 (1316.8)   & 5/5 (10460.4)  & 5/5 (39934.4)   & 0/5 (21611.0) \\
Galaxies    & 5/5 (8990.8)   & 2/5 (21186.0)  & 0/5 (17832.8)   & 0/5 (5402.2) \\
Magnets     & 5/5 (3696.6)   & 5/5 (13729.2)  & 5/5 (14493.6)   & 0/5 (3365.2) \\
Net         & 5/5 (2937.0)   & 0/5 (9123.2)   & 0/5 (13990.4)   & 0/5 (9220.2) \\
Pearl       & 5/5 (13059.0)  & 5/5 (15109.4)  & 0/5 (41637.2)   & 0/5 (20713.7) \\
Rect        & 0/5 (500.4)    & 5/5 (6514.0)   & 4/5 (14525.6)   & 0/5 (25371.6) \\
Signpost    & 3/5 (436.4)    & 1/5 (3173.8)   & 2/5 (10093.4)   & 0/5 (29715.8) \\
Singles     & 5/5 (1413.2)   & 5/5 (5431.8)   & 5/5 (9047.0)    & 1/5 (17037.2) \\
Towers      & 5/5 (1948.0)   & 5/5 (7965.2)   & 5/5 (13068.0)   & 0/5 (27154.0) \\
Tracks      & 0/5 (17736.0)  & 0/5 (4303.8)   & 0/5 (7751.8)    & 0/5 (1954.0) \\
Undead      & 5/5 (7803.0)   & 4/5 (16251.6)  & 5/5 (11390.0)   & 1/5 (25238.4) \\
Unequal     & 5/5 (2856.8)   & 5/5 (4854.8)   & 3/5 (7384.4)    & 0/5 (30060.4) \\
Unruly      & 0/5 (7246.8)   & 0/5 (10013.2)  & 0/5 (13245.4)   & 0/5 (27242.4) \\
\bottomrule
\end{tabularx}
\end{table}

\begin{table}[h]
\centering
\small
\setlength{\tabcolsep}{3pt}
\renewcommand{\arraystretch}{0.9}
\caption{\textbf{Detailed performance (Qwen3 235B).} Per-puzzle results by difficulty on the 21-puzzle suite. Each cell reports \# solved / \# evaluated instances, with mean completion token count in parentheses, computed over the evaluated instances at that difficulty. All runs use the same solver prompt and decoding settings as in the main benchmark (\cref{sec:benchmark}).}
\label{tab:puzzle_details_perf_qwen}

\begin{tabularx}{\textwidth}{YYYYYYC{2cm}C{3cm}}
\toprule
\textbf{Puzzle} & \textbf{Trivial} & \textbf{Easy} & \textbf{Normal} & \textbf{Hard}\\
\midrule
Solo        & 5/5 (5220.4)   & 0/5 (19878.2)   & 0/5 (29829.8)   & 0/5 (29961.8)\\
Tents       & 5/5 (6486.0)   & 5/5 (11591.6)   & 4/5 (21452.4)   & 0/5 (19693.3)\\
Filling     & 5/5 (8260.6)     & 0/5 (24198.3)   & 0/5 (24059.6)    & 0/5 (22894.2) \\
Lightup     & 5/5 (6770.0) & 5/5 (13369.6)& 1/5 (18392.4) & 0/5 (26117.4)\\
Mosaic      & 5/5 (6223.8)     & 5/5 (14351.0)    & 0/5 (24837.8)   & 0/5 (17945.6)\\
Keen        & 5/5 (4829.0)   & 5/5 (12328.0)   & 2/5 (24928.2)   & 0/5 (25493.4)\\
Range       & 5/5 (10886.8)     & 4/5 (20025.2)   & 0/5 (21463.8)   & 0/5 (22451.6)\\
Pattern     & 5/5 (3589.8)     & 0/ 5 (27557.2)   & 0/5 (25310.2)   & 0/5 (21853.8)\\
Dominosa    & 5/5 (4632.8)   & 3/5 (18389.8)   & 0/5 (21440.4)   & 0/5 (20329.6)\\
Galaxies    & 2/5 (17782.4)   & 0/5 (22570.0)  & 0/5 (23114.8)   & 0/5 (25084.6)\\
Magnets     & 3/5 (15607.8)  & 3/5 (28395.8)   & 2/5 (26639.0)  & 0/5 (21861.6)\\
Net         & 5/5 (10330.4)    & 0/5 (27653.0)     & 0/5 (31310.0)     & 0/5 (23810.4)\\
Pearl       & 1/5 (19109.4)   & 0/5 (20671.2)   & 0/5 (19713.0)  & 0/5 (18234.4)\\
Rect        & 0/5 (1787.8)     & 0/5 (21397.6)     & 0/5 (25064.4)     & 0/5 (22270.2)\\
Signpost    & 3/5 (2826.2)     & 1/5 (7387.6)     & 2/5 (20081.6)   & 0/5 (23929.8)\\
Singles     & 5/5 (4112.6)   & 5/5 (11495.8)   & 5/5 (13541.8)   & 0/5 (24158.975)\\
Towers      & 5/5 (9815.6)   & 5/5 (18604.0)   & 4/5 (23485.64)   & 0/5 (31988.6)\\
Tracks      & 0/5 (25576.2)   & 0/5 (20408.0)   & 0/5 (20679.6)   & 0/5 (23823.4)\\
Undead      & 1/5 (26931.2)   & 1/5 (25282.0)   & 0/5 (30382.4)   & 0/5 (24819.8)\\
Unequal     & 3/5 (6154.0)   & 5/5 (10992.4)   & 2/5 (15559.4)   & 0/5 (27065.8)\\
Unruly      & 0/5 (18470.6)   & 0/5 (23148.6)   & 0/4 (23897.6)   & 0/5 (21991.36)\\
\bottomrule
\end{tabularx}
\end{table}

\begin{table}[h]
\centering
\small
\setlength{\tabcolsep}{3pt}
\renewcommand{\arraystretch}{0.9}
\caption{\textbf{Detailed performance (Kimi K2 Thinking).} Per-puzzle results by difficulty on the 21-puzzle suite. Each cell reports \# solved / \# evaluated instances, with mean completion token count in parentheses, computed over the evaluated instances at that difficulty. All runs use the same solver prompt and decoding settings as in the main benchmark (\cref{sec:benchmark}).}
\label{tab:puzzle_details_perf_kimi}

\begin{tabularx}{\textwidth}{YYYYYYC{2cm}C{3cm}}
\toprule
\textbf{Puzzle} & \textbf{Trivial} & \textbf{Easy} & \textbf{Normal} & \textbf{Hard}\\
\midrule
Solo        & 5/5 (2302.4)   & 0/5 (45200)   & 4/5 (30893.4)   & 1/5 (56261.4)\\
Tents       & 5/5 (5734.4)   & 5/5 (16160.8)   & 5/5 (28705.2)   & 0/5 (64210.8)\\
Filling     & 5/5 (8910.4)     & 0/5 (46002.2)     & 0/5 (49713.7)    & 0/5 (60694.4)\\
Lightup     & 5/5 (2336.6) & 5/5 (12332.8) & 5/5 (31801.6) & 0/5 (65536.0)\\
Mosaic      & 5/5 (7021.2)     & 5/5 (8693.6)    & 2/5 (54524.0)   & 0/5 (59351.2)\\
Keen        & 5/5 (2894.2)   & 5/5 (13424.8)   & 4/5 (33754.4)   & 0/4 (62941.0)\\
Range       & 5/5 (5850.2)     & 4/5 (20867.0)   & 0/5 (57348.8)     & 0/5 (58553.2)\\
Pattern     & 5/5 (1438.6)     & 2/5 (55358.6)   & 0/5 (65138.0)   & 0/5 (59118.2)\\
Dominosa    & 5/5 (2134.0)   & 5/5 (30288.2)   & 0/5 (65536.0)   & 0/5 (64695.4)\\
Galaxies    & 4/5 (28160.8)   & 2/5 (56873.0)  & 0/5 (55219.6)   & 0/3 (51139.3)\\
Magnets     & 5/5 (7288.2)  & 4/5 (33953.6)   & 5/5 (34259.2)  & 0/5 (65536.0)\\
Net         & 5/5 (9660.4)    & 0/5 (25680.8)     & 0/5 (44706.2)     & 0/5 (59358.8)\\
Pearl       & 4/5 (24180.4)   & 1/5 (53557.8)   & 0/5 (62605.2)   & 0/5 (64270.6)\\
Rect        & 0/5 (1802.6)     & 2/5 (16901.0)     & 1/5 (27804.4)     & 0/5 (65536.0)\\
Signpost    & 3/5 (1662.0)     & 1/5 (8682.0)     & 1/5 (32537.4)     & 0/5 (64912.2)\\
Singles     & 5/5 (2120.4)   & 5/5 (9441.0)   & 5/5 (17775.8)   & 0/5 (65536.0)\\
Towers      & 5/5 (4798.4)   & 5/5 (17015.6)   & 5/5 (43610.4)   & 0/5 (51806.02)\\
Tracks      & 1/5 (45695.2)   & 0/5 (64871.4)   & 0/5 (62114.0)   & 0/5 (56650.4)\\
Undead      & 2/5 (35891.4)   & 4/5 (34211.8)   & 2/5 (48413.0)   & 0/5 (61080.2)\\
Unequal     & 2/5 (3252.6)   & 4/5 (22244.4)   & 4/5 (16272.8)   & 0/5 (64726.2)\\
Unruly      & 0/5 (19496.6)   & 0/5 (32227.6)   & 0/5 (56042.2)   & 0/5 (65536.0)\\
\bottomrule
\end{tabularx}
\end{table}

\begin{table}[h]
\centering
\small
\setlength{\tabcolsep}{3pt}
\renewcommand{\arraystretch}{0.9}
\caption{\textbf{Detailed performance (DeepSeek V3.2).} Per-puzzle results by difficulty on the 21-puzzle suite. Each cell reports \# solved / \# evaluated instances, with mean completion token count in parentheses, computed over the evaluated instances at that difficulty. All runs use the same solver prompt and decoding settings as in the main benchmark (\cref{sec:benchmark}).}
\label{tab:puzzle_details_perf_deepseek}

\begin{tabularx}{\textwidth}{YYYYYYC{2cm}C{3cm}}
\toprule
\textbf{Puzzle} & \textbf{Trivial} & \textbf{Easy} & \textbf{Normal} & \textbf{Hard}\\
\midrule
Solo        & 5/5 (4261.0)   & 0/5 (19392.0)   & 5/5 (39878.0)   & 0/5 (65536.0)\\
Tents       & 4/5 (3890.0)   & 5/5 (9681.6)   & 5/5 (19950.6)   & 0/5 (28922.4)\\
Filling     & 5/5 (7543.4)     & 0/5 (26743.1)     & 0/5 (25723.0)    & 0/5 (31961.8)\\
Lightup     & 5/5 (2930.4) & 5/5 (11211.2) & 5/5 (21151.2) & 0/5 (29290.4)\\
Mosaic      & 3/5 (1601.2)     & 4/5 (5567.6)    & 5/5 (21151.2)   & 0/5 (24473.2)\\
Keen        & 5/5 (3156.4)   & 5/5 (8420.2)   & 5/5 (18802.4)   & 0/5 (46462.0)\\
Range       & 5/5 (7486.8)     & 4/5 (19574.2)   & 1/5 (25182.7)     & 0/5 (33962.6)\\
Pattern     & 5/5 (2696.8)     & 2/5 (48120.0)   & 0/5 (33097.8)   & 0/5 (33757.2)\\
Dominosa    & 5/5 (3100.0)   & 4/5 (23104.8)   & 0/5 (34934.6)   & 0/5 (35696.4)\\
Galaxies    & 3/5 (18515.6)   & 3/5 (29391.4)  & 0/5 (27128.0)   & 0/5 (25761.8)\\
Magnets     & 5/5 (8124.2)  & 5/5 (25038.8)   & 4/5 (31377.8)  & 0/5 (32475.2)\\
Net         & 3/5 (5812.2)    & 0/5 (29112.6)     & 0/5 (21182.0)     & 0/5 (19381.8)\\
Pearl       & 3/5 (20979.2)   & 0/5 (35024.6)   & 0/5 (36094.2)   & 0/5 (32317.2)\\
Rect        & 0/5 (1965.8)     & 1/5 (10508.2)     & 1/5 (28211.2)     & 0/5 (36560.4)\\
Signpost    & 3/5 (1489.6)     & 0/5 (8036.6)     & 2/5 (27705.2)     & 0/5 (39522.6)\\
Singles     & 5/5 (2930.0)   & 5/5 (10367.2)   & 5/5 (14680.4)   & 0/5 (36012.4)\\
Towers      & 5/5 (7000.6)   & 5/5 (16444.8)   & 5/5 (36753.5)   & 0/5 (64782.1)\\
Tracks      & 2/5 (21102.2)   & 0/5 (31926.6)   & 0/5 (29741.8)   & 0/5 (17576.6)\\
Undead      & 5/5 (15865.4)   & 5/5 (24152.0)   & 2/5 (27259.0)   & 0/5 (36559.2)\\
Unequal     & 5/5 (3756.2)   & 3/5 (11294.8)   & 2/5 (13527.8)   & 0/5 (62481.5)\\
Unruly      &  0/5 (17382.0)   & 0/5 (30172.4)   & 0/5 (34259.1)   & 0/5 (39043.4)\\
\bottomrule
\end{tabularx}
\end{table}

\section{Prompts}
\label{app:prompts}
This appendix collects the prompts used in our pipeline: the solver prompt used to elicit puzzle solutions, the prompts used for claim extraction and screening, the prompts used for rule extraction, and the puzzle-specific claim and rule schemas.

\subsection{Solver Prompt}
\label{app:solver_prompt}
\cref{prompt:prompt_benchmark} displays the prompt used to solve the puzzles.

\begin{figure}[h]
    \centering
    \displayprompt{assets/prompt_benchmark.txt}
    \caption{Prompt for solving the instance described in the puzzle state. Puzzle descriptions are detailed in \cref{app:puzzle_descriptions}.}
    \label{prompt:prompt_benchmark}
\end{figure}

\subsection{Claim Extraction Prompts}
\label{app:claim_extraction}

This section presents the prompts used to extract the set of claims found in the model’s reasoning trace. The first prompt pair (\cref{prompt:coordinate_prompt}) determines whether the trace assumes 0-based or 1-based indexing, ensuring consistent interpretation of references. The second prompt pair (\cref{prompt:claim_extraction}) extracts atomic claims from the raw reasoning trace using a pure LLM-based approach. The third prompt pair (\cref{prompt:claim_complement}) augments the claims that are extracted by the regex by recovering missing or implicit statements that are difficult to capture with rules alone. The fourth prompt pair (\cref{prompt:claim_cleaning}) filters, normalizes, and removes redundant or ill-formed claims to produce a clean claim set. 

\begin{figure}[h]
    \centering
    \displayprompt{assets/prompt_puzzle_coordinate_base_system.txt}
    \displayprompt{assets/prompt_puzzle_coordinate_base_user.txt}
    \caption{System (top) and user (bottom) prompts for determining whether the reasoning trace relies on 0-based or 1-based indexing.}
    \label{prompt:coordinate_prompt}
\end{figure}

\begin{figure}[h]
    \centering
    \displayprompt{assets/prompt_puzzle_pure_llm_system.txt}
    \displayprompt{assets/prompt_puzzle_pure_llm_user.txt}
    \caption{System (top) and user (bottom) prompts for extracting the claims from the reasoning trace.}
    \label{prompt:claim_extraction}
\end{figure}

\begin{figure}[h]
    \centering
    \displayprompt{assets/prompt_puzzle_complement_system.txt}
    \displayprompt{assets/prompt_puzzle_complement_user.txt}
    \caption{System (top) and user (bottom) prompts for complementing claims extracted with a rule-based method.}
    \label{prompt:claim_complement}
\end{figure}

\begin{figure}[h]
    \centering
    \displayprompt{assets/prompt_puzzle_screen_system.txt}
    \displayprompt{assets/prompt_puzzle_screen_user.txt}
    \caption{System (top) and user (bottom) prompts for filtering and cleaning claims.}
    \label{prompt:claim_cleaning}
\end{figure}

\subsection{Rule Extraction Prompts}
\label{app:rule_extraction}
The rule extraction prompt (\cref{prompt:rule_extraction}) identifies deduction rules between claims, which define the directed edges of the reasoning graph (between premises and conclusions).

\begin{figure}[h]
    \centering
    \displayprompt{assets/prompt_puzzle_rule_system.txt}
    \displayprompt{assets/prompt_puzzle_rule_user.txt}
    \caption{System (top) and user (bottom) prompts for extracting deduction rules (i.e., edges).}
    \label{prompt:rule_extraction}
\end{figure}

\subsection{Puzzle-specific Claims and Rules}
\label{app:claims_rules}
These prompts are puzzle-specific and define the type of rules and claims that the model should extract for a given puzzle. The claim and rule types for \emph{Tents} are shown in \cref{prompt:claims_tents} and \cref{prompt:rules_tents}, and for Sudoku in \cref{prompt:claims_sudoku} and \cref{prompt:rules_tents}, respectively.

\begin{figure}[h]
    \centering
    \displayprompt{assets/tents_claims_description.txt}
    \caption{Possible claims for \emph{Tents}.}
    \label{prompt:claims_tents}
\end{figure}

\begin{figure}[h]
    \centering
    \displayprompt{assets/tents_rule_description.txt}
    \caption{Possible rules for \emph{Tents}.}
    \label{prompt:rules_tents}
\end{figure}

\begin{figure}[h]
    \centering
    \displayprompt{assets/sudoku_claims_description.txt}
    \caption{Possible claims for \emph{Sudoku}.}
    \label{prompt:claims_sudoku}
\end{figure}

\begin{figure}[h]
    \centering
    \displayprompt{assets/sudoku_rules_description.txt}
    \caption{Possible rules for \emph{Sudoku}.}
    \label{prompt:rules_sudoku}
\end{figure}

\newpage
\subsection{Puzzle Descriptions}
\label{app:puzzle_descriptions}

Information and details specific to each puzzle are presented in \cref{tab:puzzle_details}. Below, we provide the full textual descriptions for each puzzle used in our suite, which define the rules and expected solution format.

\begin{table}[h]
\centering
\small
\setlength{\tabcolsep}{3pt}
\renewcommand{\arraystretch}{0.9}
\caption{Puzzle size specifications by difficulty level and computational complexity.}
\label{tab:puzzle_details}

\begin{tabularx}{\textwidth}{YYYYYYC{2cm}C{3cm}}
\toprule
\textbf{Puzzle} & \textbf{Description} & \textbf{Trivial} & \textbf{Easy} & \textbf{Normal} & \textbf{Hard} & \textbf{Complexity} & \textbf{Reference} \\
\midrule
Solo        & \cref{prompt:solo}        & 4x4dt   & 9x4dt   & 9x9dt   & 9x9de      & NP-complete & \citet{yato2003complexity} \\
Tents       & \cref{prompt:tents}       & 4x4de   & 6x6de   & 8x8de   & 15x15de    & NP-complete & \citet{debiasi2012complexity} \\
Filling     & \cref{prompt:filling}     & 3x3     & 9x7     & 13x9    & 17x13      & Open & -- \\
Lightup     & \cref{prompt:lightup}     & 3x3d0s4 & 5x5d0s4 & 7x7d0s4 & 14x14d0s2  & NP-complete & \citet{mcphail2005light} \\
Mosaic      & \cref{prompt:mosaic}      & 3x3     & 5x5     & 10x10   & 25x25      & NP-complete & \citet{dejong2023mosaic} \\
Keen        & \cref{prompt:keen}        & 3x3de   & 4x4de   & 6x6dn   & 9x9dn      & Open & -- \\
Range       & \cref{prompt:range}       & 3x3     & 6x4     & 9x6     & 13x9       & Open & -- \\
Pattern     & \cref{prompt:pattern}     & 2x2     & 10x10   & 15x15   & 25x25      & NP-complete & \citet{ueda1996nonogram} \\
Dominosa    & \cref{prompt:dominosa}    & 3x2dt   & 5x4dt   & 8x7db   & 11x10db    & Open & -- \\
Galaxies    & \cref{prompt:galaxies}    & 3x3dn   & 5x5dn   & 7x7dn   & 15x15dn    & NP-complete & \citet{friedman2002spiral} \\
Magnets     & \cref{prompt:magnets}     & 3x2de   & 6x5de   & 6x5dtS  & 10x9dtS    & NP-complete & \citet{kolker2012magnets} \\
Net         & \cref{prompt:net}         & 2x2     & 4x4     & 5x5     & 11x11      & NP-complete & \citet{kral2004tough} \\
Pearl       & \cref{prompt:pearl}       & 5x5de   & 6x6de   & 8x8de   & 10x10dt    & NP-complete & \citet{friedman2002pearl} \\
Rect        & \cref{prompt:rectangles}  & 2x1     & 5x5     & 7x7     & 15x15      & NP-complete & \citet{takenaga2013shikaku} \\
Signpost    & \cref{prompt:signpost}    & 2x1     & 3x3     & 4x4     & 7x7        & NP-complete & \citet{maarse2019npcomplete} \\
Singles     & \cref{prompt:singles}     & 2x2de   & 4x4de   & 5x5de   & 12x12de    & Open & -- \\
Towers      & \cref{prompt:towers}      & 3x3de   & 4x4de   & 5x5de   & 6x6de      & NP-complete & \citet{maarse2019npcomplete} \\
Tracks      & \cref{prompt:tracks}      & 4x4de   & 8x8de   & 8x8dt   & 15x15dt    & Open & -- \\
Undead      & \cref{prompt:undead}      & 3x3de   & 4x4de   & 4x4dn   & 7x7dn      & Open & -- \\
Unequal     & \cref{prompt:unequal}     & 3x3dt   & 4x4de   & 4x4dk   & 7x7dk      & NP-complete & \citet{maarse2019npcomplete} \\
Unruly      & \cref{prompt:unruly}      & 6x6dt   & 8x6dt   & 8x8dt   & 14x14dt    & NP-complete & \citet{debiasi2012binary} \\
\bottomrule
\end{tabularx}
\end{table}

\begin{figure}[h]
    \centering
    \displayprompt{assets/puzzle_descriptions/dominosa.txt}
    \caption{Puzzle description for \emph{Dominosa}.}
    \label{prompt:dominosa}
\end{figure}

\begin{figure}[h]
    \centering
    \displayprompt{assets/puzzle_descriptions/filling.txt}
    \caption{Puzzle description for \emph{Filling}.}
    \label{prompt:filling}
\end{figure}

\begin{figure}[h]
    \centering
    \displayprompt{assets/puzzle_descriptions/galaxies.txt}
    \caption{Puzzle description for \emph{Galaxies}.}
    \label{prompt:galaxies}
\end{figure}

\begin{figure}[h]
    \centering
    \displayprompt{assets/puzzle_descriptions/keen.txt}
    \caption{Puzzle description for \emph{Keen}.}
    \label{prompt:keen}
\end{figure}

\begin{figure}[h]
    \centering
    \displayprompt{assets/puzzle_descriptions/lightup.txt}
    \caption{Puzzle description for \emph{Lightup}.}
    \label{prompt:lightup}
\end{figure}

\begin{figure}[h]
    \centering
    \displayprompt{assets/puzzle_descriptions/magnets.txt}
    \caption{Puzzle description for \emph{Magnets}.}
    \label{prompt:magnets}
\end{figure}

\begin{figure}[h]
    \centering
    \displayprompt{assets/puzzle_descriptions/mosaic.txt}
    \caption{Puzzle description for \emph{Mosaic}.}
    \label{prompt:mosaic}
\end{figure}

\begin{figure}[h]
    \centering
    \displayprompt{assets/puzzle_descriptions/net.txt}
    \caption{Puzzle description for \emph{Net}.}
    \label{prompt:net}
\end{figure}

\begin{figure}[h]
    \centering
    \displayprompt{assets/puzzle_descriptions/pattern.txt}
    \caption{Puzzle description for \emph{Pattern}.}
    \label{prompt:pattern}
\end{figure}

\begin{figure}[h]
    \centering
    \displayprompt{assets/puzzle_descriptions/pearl.txt}
    \caption{Puzzle description for \emph{Pearl}.}
    \label{prompt:pearl}
\end{figure}

\begin{figure}[h]
    \centering
    \displayprompt{assets/puzzle_descriptions/range.txt}
    \caption{Puzzle description for \emph{Range}.}
    \label{prompt:range}
\end{figure}

\begin{figure}[h]
    \centering
    \displayprompt{assets/puzzle_descriptions/rectangles.txt}
    \caption{Puzzle description for \emph{Rectangles}.}
    \label{prompt:rectangles}
\end{figure}

\begin{figure}[h]
    \centering
    \displayprompt{assets/puzzle_descriptions/signpost.txt}
    \caption{Puzzle description for \emph{Signpost}.}
    \label{prompt:signpost}
\end{figure}

\begin{figure}[h]
    \centering
    \displayprompt{assets/puzzle_descriptions/singles.txt}
    \caption{Puzzle description for \emph{Singles}.}
    \label{prompt:singles}
\end{figure}

\begin{figure}[h]
    \centering
    \displayprompt{assets/puzzle_descriptions/solo.txt}
    \caption{Puzzle description for \emph{Solo} (also known as \emph{Sudoku}).}
    \label{prompt:solo}
\end{figure}

\begin{figure}[h]
    \centering
    \displayprompt{assets/puzzle_descriptions/tents.txt}
    \caption{Puzzle description for \emph{Tents}.}
    \label{prompt:tents}
\end{figure}

\begin{figure}[h]
    \centering
    \displayprompt{assets/puzzle_descriptions/towers.txt}
    \caption{Puzzle description for \emph{Towers}.}
    \label{prompt:towers}
\end{figure}

\begin{figure}[h]
    \centering
    \displayprompt{assets/puzzle_descriptions/tracks.txt}
    \caption{Puzzle description for \emph{Tracks}.}
    \label{prompt:tracks}
\end{figure}

\begin{figure}[h]
    \centering
    \displayprompt{assets/puzzle_descriptions/undead.txt}
    \caption{Puzzle description for \emph{Undead}.}
    \label{prompt:undead}
\end{figure}

\begin{figure}[h]
    \centering
    \displayprompt{assets/puzzle_descriptions/unequal.txt}
    \caption{Puzzle description for \emph{Unequal}.}
    \label{prompt:unequal}
\end{figure}

\begin{figure}[h]
    \centering
    \displayprompt{assets/puzzle_descriptions/unruly.txt}
    \caption{Puzzle description for \emph{Unruly}.}
    \label{prompt:unruly}
\end{figure}

\end{document}